\documentclass[a4paper,12pt]{elsarticle}
\usepackage{hyperref}
\usepackage{natbib}

\usepackage{algorithm,algorithmic,amssymb,amsmath,letltxmacro}
\newlength{\commentindent}
\makeatletter
\renewcommand{\algorithmiccomment}[1]{\unskip\hfill\makebox[\commentindent][l]{//~#1}\par}
\LetLtxMacro{\oldalgorithmic}{\algorithmic}
\renewcommand{\algorithmic}[1][0]{%
  \oldalgorithmic[#1]%
  \renewcommand{\ALC@com}[1]{%
    \ifnum\pdfstrcmp{##1}{default}=0\else\algorithmiccomment{##1}\fi}%
}
\makeatother

\biboptions{sort&compress}   

\newfont{\shit}{msbm10 scaled 1200}
\newfont{\shiti}{msbm7 scaled 1200}
\newcommand{\IS}{\mbox{\shit{S}}}

\newcommand{\xibf}{\mbox{\boldmath{$\xi$}}}
\newcommand{\phibf}{\mbox{\boldmath{$\phi$}}}
\newcommand{\betabf}{\mbox{\boldmath{$\beta$}}}
\newcommand{\omegabf}{\mbox{\boldmath{$\omega$}}}

\journal{Journal of the Mechanics and Physics of Solids}

\fontfamily{ptm} \selectfont

\renewcommand{\baselinestretch}{1.3}
\usepackage{bm}
\usepackage{cancel}
\usepackage{amsmath}
\usepackage{amssymb}
\usepackage{url}
\usepackage{booktabs}
\usepackage{bbm}
\usepackage{upgreek}

\newcommand{\Is}{\mbox{\shiti{S}}}

\newcommand{\bbf}{\mbox{\bf b}}
\newcommand{\Cbf}{\mbox{\bf C}}
\newcommand{\Ebf}{\mbox{\bf E}}
\newcommand{\Fbf}{\mbox{\bf F}}
\newcommand{\Ibf}{\mbox{\bf I}}
\newcommand{\Nbf}{\mbox{\bf N}}

\newcommand{\fbf}{\mbox{\bf f}}
\newcommand{\gbf}{\mbox{\bf g}}

\newcommand{\xbf}{\mbox{\bf x}}

\newcommand{\wbf}{\mbox{\bf w}}
\newcommand{\zbf}{\mbox{\bf z}}
\newcommand{\Xbf}{\mbox{\bf X}}
\newcommand{\sigbf}{\mbox{\boldmath{$\sigma$}}}

\newcommand{\vv}[1]{{\boldsymbol #1}}
\newcommand{\p}[2]{{p}(#1\mid #2)}
\newcommand{\prior}[1]{{p}(#1)}

\newcommand{\expect}[1]{\langle #1 \rangle}
\newcommand{\xiset}{\vv \Xi}

\newcommand{\obs}{\sigma}
\newcommand{\obssmallset}{\Sigma^{(n_{\xi})}}
\newcommand{\obsbigset}{\vv \Sigma}

\newcommand{\surryesno}{}

\begin{document}

\begin{center}
	{\bf \Large Stochastic Modeling of Inhomogeneities in the Aortic Wall \\[-1.2ex] and Uncertainty Quantification using a\\[0.1ex] Bayesian Encoder-Decoder Surrogate}
	
	\vspace*{0.25cm}
	
\renewcommand{\thefootnote}{\fnsymbol{footnote}}
Sascha Ranftl$^{a,b,}$\footnote[2]{Contributed equally}, Malte Rolf-Pissarczyk$^{c,\dagger}$, Gloria Wolkerstorfer$^a$, Antonio Pepe$^d$, \\[-0.6ex] Jan Egger$^e$, Wolfgang von der Linden$^a$, Gerhard A. Holzapfel$^{c,f,}$\footnote[1]{Corresponding author: holzapfel@tugraz.at}

	\vspace*{0.25cm}
	
	$^a$Graz University of Technology, Institute of Theoretical and Computational Physics, Austria
	
	$^b$Graz Center for Computational Engineering, Graz University of Technology, Austria
	
	$^c$Graz University of Technology, Institute of Biomechanics, Austria
	
	$^d$Graz University of Technology, Institute of Computer Graphics and Vision, Austria

	$^e$University Medicine Essen, Institute for AI in Medicine, Essen, Germany

	$^f$Norwegian University of Science and Technology (NTNU)\\[-1.1ex] Department of Structural Analysis, Trondheim, Norway\\[-0.0ex]
	
	\vspace*{0.4cm}
	
	{Submitted   \bf { }}\\[-0.85ex]
	February 21, 2022
	
\end{center}

\renewcommand{\baselinestretch}{1.0}
\small\normalsize

\noindent {\bf Abstract.}
Inhomogeneities in the aortic wall can lead to localized stress accumulations, possibly initiating dissection. 
In many cases, a dissection results from pathological changes such as fragmentation or loss of elastic fibers. But it has been shown that even the healthy aortic wall has an inherent heterogeneous microstructure. 
Some parts of the aorta are particularly susceptible to the development of inhomogeneities due to pathological changes, however, the distribution in the aortic wall and the spatial extent, such as size, shape, and type, are difficult to predict.
Motivated by this observation, we describe the heterogeneous distribution of elastic fiber degradation in the dissected aortic wall using a stochastic constitutive model. 
For this purpose, random field realizations, which model the stochastic distribution of degraded elastic fibers, are generated over a non-equidistant grid.
The random field then serves as input for a uniaxial extension test of the pathological aortic wall, solved with the finite-element (FE) method. To include the microstructure of the dissected aortic wall, a constitutive model developed in a previous study is applied, which also includes an approach to model the degradation of interlamellar elastic fibers.
Then to assess the uncertainty in the output stress distribution due to this stochastic constitutive model, a convolutional neural network, specifically a Bayesian encoder-decoder, was used as a surrogate model that maps the random input fields to the output stress distribution obtained from the FE analysis.
The results show that the neural network is able to predict the stress distribution of the FE analysis while significantly reducing the computational time.
In addition, it provides the probability for exceeding critical stresses within the aortic wall, which could allow for the prediction of delamination or fatal rupture.

\vspace{0.15cm}

\noindent {\bf Keywords.}
Stochastic constitutive modeling; finite-element analysis; fibrous tissue; aortic dissection; beta random field; Bayesian encoder-decoder; uncertainty quantification


\renewcommand{\baselinestretch}{1.3}
\small\normalsize






\section{Introduction}\label{s:1}
%
Material inhomogeneities can have a considerable influence on the mechanical behavior of the aorta. When investigating healthy aortas, spatial and temporal variations in the mechanical behavior are often observed, which can be explained by different factors such as age, lifestyle or gender \cite{Astrand1985a,Stefanadis1997a,Roccabianca2014c}. 
Observations range from sharp strain localization under uniaxial loading \cite{DiGiuseppe2021a} to large variations in the stress-stretch ratio of tissue samples taken from adjacent sites under uniaxial or biaxial loading \cite{Roccabianca2014c}. In particular, inhomogeneities may be associated with pathological changes in the aortic wall, which are often remodeling processes that are triggered by hemodynamic changes \cite{Humphrey2015a}, in particular by changed wall shear stresses. 
Pathological changes are usually preceded by altered stress distributions in the aortic wall or inflammatory processes \cite{Halushka2014a}. This then leads to spatial and temporal variations in the fraction of individual aortic constituents, such as collagen fibers, elastic fibers, smooth muscle cells or the ground substance \cite{Tsamis2013a,Lacolley2018a}. 

Material inhomogeneities play a major role in the initiation and progression of an aortic dissection. Defects in the microstructure of the aortic wall can potentially lead to localized stress accumulations that induce aortic wall delamination. In particular, glycosaminoglycans (GAGs) may play an important role in the pathology of aortic dissection, as hypothesized by Humphrey~\cite{Humphrey2013a}. The accumulation of GAGs between the elastic lamellae may cause a swelling pressure that leads to separation of the elastic lamellae, which can lead to delamination of the aortic wall. However, not only local accumulation of GAGs is associated with aortic dissection, but also apoptosis and smooth muscle cell dyfunction, remodeling of collagen fibers, and fragmentation or loss of elastic fibers \cite{Humphrey2013a}.
However, these pathological alterations do not occur individually. Their onset is often correlated, which might be due to the highly interconnected microstructure of the aorta. For example, the swelling pressure caused by pooled GAGs results in degradation of interlamellar, radially-oriented elastic fibers. Elastic fibers, in turn, connect smooth muscle cells to the elastic lamellae. As a result of elastic fiber degradation, smooth muscle cells lose their microstructural integrity, leading to apoptosis and dysfunction \cite{Shen2017a}. However, the exact mechanisms are not yet fully understood.

\begin{figure}[t]
    \centering   \hspace*{-0.5cm}  
    \includegraphics[scale=0.82]{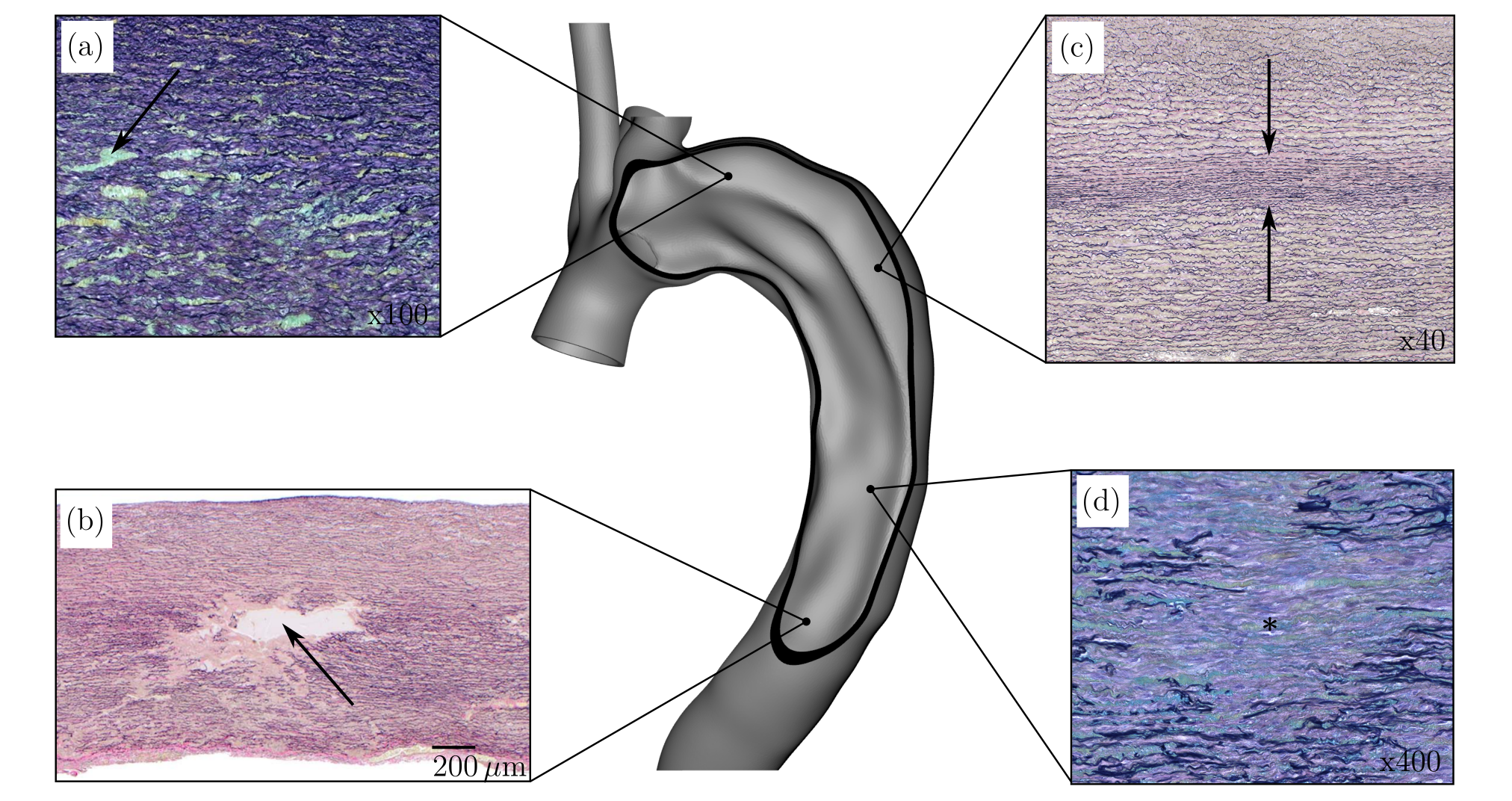}
    \vspace*{-0.75cm}
    \caption{
    Cut through a patient-specific geometry of an aortic dissection with some related pathological changes of the microstructure, particularly the accumulation of mucoid matrix material between the elastic lamellae and the fragmentation or loss of elastic fibers.
    Depending on the location along the aorta, the aortic wall can undergo different pathological changes of varying degrees. For example, studies have linked the accumulation of mucoid matrix material, or GAGs, between the elastic lamellae of different sizes (arrow) (a), (b), the band-like collapse of elastic lamellae (arrows) (c), and the fragmentation or loss of elastic fibers in a patchy fashion (asterisk) (d), to the pathology of aortic dissections. Note that thin (c) and thick (d) lines indicate elastic lamellae.  
    Histological images reprinted from Eleid et al.~\cite{Eleid2013a} and Halushka et al.~\cite{Halushka2016a}.    
    }
    \label{fig:aorticdissection}
    \vspace*{-0.1cm}
\end{figure}

As presented in Fig.~\ref{fig:aorticdissection}, histological studies show  that pathological alterations described above are more localized. Their shape and size also vary significantly. While pooled GAGs often occur in spherical shapes of different sizes \cite{Borges2009a,Cikach2018a}, as shown in Figs.~\ref{fig:aorticdissection}(a) and (b), apoptosis and smooth muscle cell dysfunction are usually found incomplete or a in band-like fashion \cite{Halushka2016a}. Similarly, studies have demonstrated a band-like collapse of elastic lamellae and patchy loss of elastic fibers in the aortic wall, as illustrated in Figs.\ref{fig:aorticdissection}(c) and (d), respectively.
A connection between apoptosis and dysfunction of smooth muscle cells and the degradation of elastic fibers cannot therefore be excluded. The influence of these local phenomena on the stress distribution is still unclear and challenging to model, since experiments do not yet provide reliable data on the global distribution of pathological changes in the aorta. In most cases, histological or immunohistochemical methods only focus on local phenomena. However, as shown in computational studies, these changes have the potential to significantly affect the local stress distribution in the aortic tissue.

In regard to aortic dissection, there are only a few computational studies examining local inhomogeneities in the aortic wall~\cite{Pepe2020a}. In two consecutive studies, Roccabianca et al.~\cite{Roccabianca2014a,Roccabianca2014b} investigated the effect of local inhomogeneities in the dissected aorta by incorporating an inclusion of GAGs into a rectangular slap of the media. The size and shape of inclusion of GAGs were based on experiments.
In addition, the authors assigned only a low tensile stiffness to the inclusion, and osmotic loading of the media and inclusion was built in. After implementing the model in a finite-element (FE) analysis software, circumferential and axial stresses were locally increased around the
periphery of the inclusion. Based of these findings, later, Ahmadzadeh et al.~\cite{Ahmadzadeh2018a, Ahmadzadeh2019a} later used a smooth particle approach to study the potential role of pooled GAGs in the initiation and progression of intralamellar delamination by modeling the coalescence and growth of GAGs in the dissected aortic wall. The computational results showed significant intramural stress concentrations and a transition from normal compressive to unusual tensile stress in the radial direction near the tip of a GAG pool, consistent with the results of Roccabianca et al.~\cite{Roccabianca2014a,Roccabianca2014b}. 
Both approaches provide evidence that local inhomogeneities, such as the accumulation of GAGs, can lead to local stress concentration, potentially leading to intralamellar delamination. However, for reasons of computational efficiency, these model approaches are not appropriate to investigate more general boundary-value problems related to healthy or dissected aortas. Also, the importance of size, shape, and location in the aorta has never been modeled or studied.

There are practically no experimental studies on the spatial and temporal distribution of material inhomogeneities over large sections of the aorta, neither in the healthy nor in the diseased case. Therefore, a stochastic model approach is necessary to investigate the effects of heterogeneities in the aortic wall on the stress distribution. 
In particular, the application of random fields \cite{Biehler2015a,Staber2018a} is an obvious possibility.

The lack of experimental data about the inherent physiological variability and additionally the limited knowledge about, e.g., {\em in vivo} boundary conditions leads to uncertainties in the computational model that  are difficult to quantify. 
By treating the model input as stochastic, the outcome is automatically uncertain as well. In this work, specific material parameters of a computational model are described by a random field. As a result, the computational model can no longer be considered deterministic and is therefore quite meaningless without propagating the uncertainties through the model. Instead, a probabilistic description of the uncertain model parameters is required, resulting in a stochastic problem that needs to be defined and solved. 

Random fields allow a material parameter to be treated as a collection of random variables at each point in the domain, meaning that the parameter takes on a stochastic value and a different such value at each point in the domain. In real aortic tissue, adjacent points are not statistically independent. If the parameter takes a certain value at a point,
then the parameter should statistically have similar albeit not identical values at adjacent points. This similarity can be described and controlled by the correlation between any two points. Besides that, random fields can also be constructed to obey certain statistics, e.g., a Gaussian or uniform distribution. If the modeled parameter is known to be within a fixed range over the entire domain, a beta distribution or a uniform distribution is a reasonable choice. Random fields offer a compelling way to model a material parameter with all of these properties; stochastic, spatially inhomogeneous, spatially correlated and possibly bounded. In this work, we will construct a random field to model a material parameter that has all these properties.

Apart from random fields, another stochastic approach to modeling heterogeneous materials is based on stochastic volume elements and homogenization. This multiscale approach involves statistical volume elements of the microstructure to obtain the macroscopic and homogenized response of a material by volumetric averaging \cite{Clement2012a,Ma2015a}, which builds on the general concept of representative volume elements describing the unit cell in a periodic microstructure \cite{Dalbosco2021a,Dalbosco2022a}. Random fields offer a convincing alternative if the macroscale is to be modeled heterogeneously and the stochastic properties of the random field can be derived from experimental data \cite{DiGiuseppe2021a} or the statistics of microscale simulations, e.g., with statistical volume elements \cite{Jeulin2001a,Jeulin2004a,OstojaStarzewski2006a,Sanei2015a}. 
This is particularly appealing because random fields then allow, in principle, to identify a parametrized form of the spatial correlation structure of a material \cite{Rasmussen2006a}, and so a physical law that can then be used for predictions. This is in contrast to purely statistical volume elements, where a non-parametric spatially correlated structure can be estimated \cite{Zhang2013a,Zhang2014a}. Random fields and statistical volume elements are therefore complementary methods \citep{Jeulin2001a}. In this work, we focus on modeling the macroscopic scale with random fields alone. Random fields enable the inclusion of prior information and can seamlessly and consistently be integrated into the uncertainty propagation by applying Bayesian probability theory \cite{vonderLinden2014a,Ranftl2021b}. Since much of the random field theory is based on stochastic processes, especially Gaussian processes, a large amount of advanced mathematics is available.

While stochastic constitutive modeling has recently attracted increasing attention, so has the problem of quantifying uncertainties of model output, particularly in the field of biomedical engineering \cite{Eck2016a}. The aim of this study is therefore to model inhomogeneities in the dissected aortic wall using a stochastic approach. More specifically, a beta random field with the special case of a uniform distribution is used to describe the stochastic, spatially distributed, and spatially correlated degradation of interlamellar elastic fibers by a degradation parameter based on the previous constraint that the degradation parameter is bounded. Due to the treatment of the constitutive model input as stochastic, the outcome of the FE analysis, more precisely the stress distribution, is also subject to uncertainties. It is therefore important to quantify these uncertainties \cite{Ghanem2017a}. This endeavor implies the need to perform FE simulations for a large number of random field realizations. Since this is not computationally feasible for FE models with many degrees of freedom, a surrogate model is used. 

Traditional surrogate models for uncertainty quantification (UQ), such as polynomial chaos expansions \cite{Xiu2005a,Crestaux2009a} or Gaussian processes regression \cite{OHagan1978a,Rasmussen2006a} are difficult to construct for the dependent variables of random fields. They also suffer from the curse of dimensionality, i.e. they scale unfavorably with the number of stochastic variables. This severely limits the spatial resolution of the stochastic constitutive model. In contrast, neural networks (NN), used as a surrogate model, do not suffer from this particular short-coming and practically increase the possibilities for uncertainty propagation of stochastic models. In addition, deep and convolutional neural networks (CNN) have proven to be able to capture spatial correlation information \cite{Tripathy2018a}. Therefore, in this study, a NN is trained on a comparatively small number of FE simulations and used as a surrogate model. To perform UQ of the stress distributions, a Bayesian deep convolutional encoder-decoder, as proposed by Zhu et al.~\cite{Zhu2018a}, is applied, which is known to learn information about spatial correlation within data. 
We use the property that both the input random field and the output stress distribution have the same grid-like structure. 
Once the NN has learned the simulation output as a function of the random field, it can approximately predict the outcome of the FE simulations with drastically reduced computation time. In other related literature \cite{Liang2018a} it is shown that it is possible to learn an accurate deep learning-based surrogate for the stress distribution on realistic domains of the aorta as a function of geometry, but not for a constitutive model. Liu et al.~\cite{Liu2021a} used a machine learning-based surrogate model to directly compute an aortic wall failure metric, again as a function of geometrical parameters and for homogeneous material parameters. Differently, He et al.~\cite{He2021a} estimated the risk of thoracic aortic aneurysm rupture using different machine learning models trained with experimental {\em in vitro} data. The combination of the experimental data-based approach in \cite{He2021a} with the probabilistic modeling in this study in conjunction with Bayesian model comparison \cite{Sivia2006a} could potentially lead to the identification of a parametrized, stochastic model for the spatial correlation structure of tissue inhomogeneities.

The present study is structured as follows. In Section~\ref{s:2}, the stochastic constitutive model framework is introduced. In this context, the constitutive model of the aorta is introduced, which describes the behavior of the aortic constituents including the degradation of interlamellar elastic fibers by introducing a degradation parameter. Next, a beta random field of the spatially distributed degradation parameter is generated, sampled and then applied to a uniaxial extension test of the aortic wall, which is solved with the FE method. To quantify the uncertainties in the computational model, Section~\ref{s:3} then sketches a convolutional network as a surrogate model, in particular a Bayesian encoder-decoder, that learns to predict the stress distribution of the FE analysis using the random field as input. Subsequently, the uncertainties of the NN are assessed, and the results obtained from the UQ of the computational model are presented in Section~\ref{s:4}. Finally, Section~\ref{s:5} summarizes and discusses the study presented, also with a view to future work. 

\section{Stochastic constitutive model framework}\label{s:2}
In this section, the constitutive model framework is presented in a stochastic manner. The constitutive model describes the behavior of the aortic wall by explicitly including its constituents, namely collagen fibers, elastic fibers and the ground substance. In addition, it models the degradation of interlamellar elastic fibers during an aortic dissection. A stochastic description of the degradation parameter is then attempted using a random field \cite{Guilleminot2011a,Staber2018a}. To do this, we construct the degradation parameter as a beta random field from an auxiliary Gaussian random field. In fact, we generate samples of Gaussian random fields and then map them to beta random fields. Finally, the entire framework is applied to a boundary-value problem.

\subsection{Constitutive model framework}\label{ss:2.1}
%
In the healthy media, elastic laminae are interconnected by interlamellar elastic fibers that are primarily radially oriented. As previously discussed, there are several pathological findings associated with aortic dissection, one of which is the local accumulation of GAGs in the media which leads to a swelling pressure between the elastic laminae. This swelling pressure is related to the rupture of elastic fibers in the radial direction, which correlates with the observation that interlamellar elastic fibers are often degraded in aortic dissection, as shown in Fig.~\ref{fig:aorticdissection}. The approach of Rolf-Pissarczyk et al.~\cite{RolfPissarczyk2021a}, as recapitulated below, assumes that elastic laminae and interlamellar elastic fibers can be accounted for by a dispersion of elastic fibers, while interlamellar elastic fibers are assumed to be symmetrically dispersed in the lamellar unit of the media. Diseased or degraded elastic fibers are then excluded. 

We first introduce the deformation gradient $\Fbf$ relative to a predefined reference configuration. If we consider an incompressible material, we require that the determinant of $\Fbf$, known as the Jacobian $J$, is equal to unity or $\mathrm{det}\Fbf\equiv 1$. For this model we can now decouple $\Fbf$ into a volumetric (dilatational) part $J^{1/3}\Ibf$ and an isochoric (distortional) part $\overline{\Fbf} = J^{-1/3}\Fbf$, where $\Ibf$ is the second-order unit tensor. The right Cauchy–Green tensor $\Cbf = \Fbf^{\rm T}\Fbf$ is the basic kinematic variable formulated in the reference configuration, together with its modified counterpart $\overline{\Cbf} = \overline{\Fbf}^{\rm T} \overline{\Fbf}$ and the corresponding first invariants $I_1 = \mathrm{tr}\,\Cbf$ and $\bar{I}_1=\mathrm{tr}\,\overline{\Cbf}$. 

The direction of a fiber in the reference configuration, denoted by the vector $\Nbf$, is given by 
\begin{equation}
	\Nbf (\Theta,\Phi) = \sin \Theta\cos \Phi\Ebf_1 + \sin\Theta\sin \Phi\Ebf_2 + \cos\Theta\Ebf_3, 
\end{equation}
where $\Ebf_i$, $i=1,2,3$, are the Cartesian unit basis vectors, while $\Theta$ and $\Phi$ are the polar and azimuth angles, respectively. We further define that the unit vector $\Nbf(\Theta)$ lies on the unit hemisphere $\IS = \{(\Theta, \Phi) | \Theta \in [0, \pi], \Phi \in [0, \pi]\}$. Because of symmetry, only half of the unit hemisphere needs to be considered. Then we discretize the unit hemisphere into a finite number of elementary areas $\Delta \IS_n$, $n = 1,\ldots,m$, more precisely spherical triangles.

By assuming a hyperelastic material, we now introduce the strain-energy function $\Psi$ in a decoupled form as 
\begin{equation}
	\Psi = \Psi_{\rm vol} + \Psi_{\rm iso},
\end{equation}
where $\Psi_{\rm vol}$ and $\Psi_{\rm iso}$ represent the purely volumetric and isochoric parts, respectively \cite{Holzapfel2000b}. The volumetric part can be defined as
\begin{equation}
    \Psi_{\rm vol} = \frac{K}{4}(J^2-1-2\ln{J}),
\end{equation}
and the isochoric part can be further decomposed into
\begin{equation}
   \Psi_{\rm iso} = \Psi_{\rm g} + \Psi_{\rm c} + \Psi_{\rm e},
\end{equation}
where $\Psi_{\rm g}$ represents the ground substance modeled by a neo-Hookean model, and $\Psi_{\rm c}$ and $\Psi_{{\rm e}}$ represent the energies stored in the collagen and elastic fibers, respectively.

To formulate the strain-energy function of elastic fibers in terms of the discrete fiber dispersion (DFD) method \cite{Li2018a}, we can write
\begin{equation}\label{eq:DFD_ef}
	\Psi_{\rm e} = \sum_{n=1}^m \rho_{{\rm e}n} 
	\Psi_{{\rm e}n}(\bar{I}_{4{\rm e}n}), 
\end{equation}
where $\rho_{{\rm e}n}$ defines the discrete density of a fiber,  $\Psi_{{\rm e}n}(\bar{I}_{4n})$ is the single fiber strain energy that is given by a general fiber model \cite{Markert2005a}, and $\bar{I}_{4{\rm e}n} = \overline{\Cbf} \colon \Nbf_n \otimes \Nbf_n$.
The choice of (\ref{eq:DFD_ef}) must ensure the condition $\Psi_n(1) = \Psi_n^\prime(1) = 0$.  After discretizing the unit hemisphere in $m$ elementary areas, the discrete density $\rho_{{\rm e}n}$ of elastic fibers can be expressed as
\begin{equation}
	\rho_{{\rm e}n} = \frac{1}{2\pi} \int_{\Delta\Is_n}
	\rho_{\rm e}(\Theta,\Phi)\sin\Theta{\rm d}\Theta{\rm d}\Phi, 
	\qquad 
 	n=1, \ldots, m.
\end{equation} 
In addition, we must satisfy the normalization condition, which by definition is satisfied by the choice of the distribution function. For the discrete approach, i.e.
\begin{equation}
    \sum_{n=1}^m \rho_{{\rm e}n} = 1.
\end{equation}

As proposed by Rolf-Pissarczyk et al.~\cite{RolfPissarczyk2021a}, we now introduce a degradation parameter $\xi$ to describe the degradation of elastic fibers as a result of separated elastic laminae, 
\begin{equation} \label{eq:degradation-parameter}
    \xi\in[0,1],
\end{equation}
where $\xi=0$ is associated with a healthy tissue and $\xi=1$ with a completely diseased (damaged/degraded) tissue, which is analogous to the continuum damage theory \cite{Holzapfel2000b}. 
Then, to exclude degraded elastic fibers from the total strain-energy function, a degradation or critical fiber angle is defined as $\Theta_\xi = \pi\xi/2$. Therefore, we distinguish the cases
\newcommand{\Einschub}{
	\begin{tabular}[c]{ll}
	$f_{{\rm e}n}(\bar{I}_{4{\rm e}n})$ & $\mathrm{if} \quad \Theta_n \geq \Theta_{\xi} \quad \mbox{and} \quad I_{4{\rm e}n} \geq 1$, \\
    $0$  & $\mbox{else}$,
\end{tabular}}
\begin{equation}\label{eq:cases_degradation}
	\Psi_{{\rm e}n} = \left\{ \Einschub \right. 
\end{equation}
where $f_{\rm e}$ represents the mathematical expression of the strain-energy function of a single elastic fiber, while $I_{4{\rm e}n}=\Cbf\colon\Nbf_n\otimes\Nbf_n$. Since the degradation of elastic fibers initiates from the radial direction due to the highest stretch occurring and leads to a higher rupture vulnerability, radially-oriented elastic fibers are initially excluded. This results to a reduced delamination strength.

The isochoric part of the strain-energy function then reads
\begin{equation}\label{eq:isochoric_psi}
	\Psi_{\rm iso} = \Psi_{\rm g}(\bar{I}_1) + 
	\sum_{n=1}^m \rho_{{\rm c}n}\Psi_{{\rm c}n}
	(\bar{I}_{4{\rm c}n}) + 
	\sum_{n=1}^m \rho_{{\rm e}n}\Psi_{{\rm e}n}(\bar{I}_{4{\rm e}n}),
\end{equation}
where the strain-energy function of collagen fibers is formulated analogously to (\ref{eq:DFD_ef}), i.e. within the framework of the DFD method by using an exponential approach to model the stiffening of collagen fibers \cite{Holzapfel2000a}. In order to implement the constitutive model framework in a FE analysis software, the Cauchy stress tensor and the elasticity tensor need to be formulated. In this context, reference is made to the study of Rolf-Pissarczyk et al.~\cite{RolfPissarczyk2021a}. 

\subsection{Gaussian random fields}\label{ss:2.2}
A random field $\mathcal{F}(\Omega_0)$ is a function that takes on a random value at every point in the reference domain $\Xbf \in \Omega_0, \Xbf = (X_1, X_2, X_3)^{\rm T}$. It is also sometimes thought of as a stochastic process, but the coordinates are usually spatial and continuous rather than temporal and discrete. It can be understood colloquially as the generalization of a multivariate random variable, which is a finite collection of random variables, to an infinite collection of random variables.
A random field is said to be Gaussian if every marginal probability distribution is Gaussian, i.e. every finite subset of the infinite collection of random variables follows a (multivariate) Gaussian distribution.
A Gaussian random field can be completely described by its mean function ${{\mu}(\Xbf)}$ and a positive semi-definite covariance function $\mathrm{k}(\Xbf, \Xbf^\prime)$ between any two points $\Xbf \in \Omega_0$ and $\Xbf^\prime \in \Omega_0$. 

We now define an auxiliary Gaussian random field $\mathcal{F}$, from which a beta random field for the degradation $\xi$, as introduced in (\ref{eq:degradation-parameter}), is later constructed, i.e.
\begin{equation} \label{eq:def:gp}
    \mathcal{F} \sim {\cal{GP}}\big( {{\mu}(\Xbf}), \mathrm{k}\big(\Xbf,\Xbf^\prime ) \big),
\end{equation}
meaning that $\mathcal{F}$ is distributed according to a Gaussian process $\mathcal{GP}$. Assuming a smooth Gaussian random field that the magnitude of the local fluctuations of the degradation parameter $\xi$ between two \lq neighboring' points in the domain should be small, a suitable covariance function is the so-called squared-exponential
\begin{equation}\label{eq:def-kernel}
    \mathrm{k}(\Xbf, \Xbf^\prime) = \varsigma^2 \exp{\Big(-\frac{(\Xbf - \Xbf^\prime)^2}{2\iota^2} \Big)},
\end{equation}
where $\varsigma^2$ and $\iota$ are model parameters that define the magnitude of the variation and the length scale of the correlation, respectively. If $\Xbf = \Xbf^\prime$ then $\varsigma^2$ is the variance. The value of $\iota$ specifies the neighborhood. In the following we denote  $\iota$ as the correlation length. Note that the smoothness is defined by the shape of the covariance function, hence the choice of the covariance function significantly affects the spatial correlation structure and behavior of the random field realizations. One may prefer to choose a different covariance function, e.g., from the more general family of Mat{\'e}rn covariance functions with a smoothness parameter \cite{Rasmussen2006a}.
On the other hand, the mean function $\mu$ is not of particular importance at this point, since the Gaussian random field is later transformed and re-scaled into a beta random field. Hence, we can set $\mu \equiv 0$ without loss of generality. 
Note that this model is stationary, i.e. the value of the covariance function only depends on the distance between any two points, $\Xbf - \Xbf^\prime$, and does not change upon translation of the points in space. We will later exploit this fact for a fast computation of random field realizations. Non-stationary models usually require more advanced methods \cite{Fuglstad2015a, Fuglstad2015b}.
Moreover, the presented model can easily be extended to spatially anisotropic correlation structures by defining corresponding anisotropic covariance functions of $\Xbf = (X_1, X_2, X_3)^{\rm T}$, e.g., $\mathrm{k}(\Xbf, \Xbf^\prime) = \mathrm{k}_1(X_1, X_1^\prime)\mathrm{k}_2(X_2, X_2^\prime)\mathrm{k}_3(X_3, X_3^\prime)$.

Let $\hat{\Xbf} = \{\Xbf^{(n_{\rm x})} \}_{n_{\rm x}=1}^{N_{\rm x}}$, where $\Xbf^{(n_{\rm x})} = (X_1^{(n_{\rm x})}, X_2^{(n_{\rm x})}, X_3^{(n_{\rm x})})^{\rm T}$, be now the set of coordinates of the nodes for the discretization of the computational domain $\Omega_0$, i.e. the reference configuration. Then, $\fbf(\hat{\Xbf})$ is a finite subset of the infinite collection of random variables implied by $\mathcal{F}(\Omega_0)$, $\hat{\Xbf} \subset \Omega_0$. Note the difference between $\mathcal{F}(\Omega_0)$ as an infinite collection of random variables at all locations in the domain $\Xbf \in \Omega_0$, and $\fbf(\hat{\Xbf})$ as a finite collection of random variables at the nodes of the domain discretization $\Xbf \in \hat{\Xbf}, \hat{\Xbf} \in \Omega_0$. Both the random field $\mathcal{F}(\Omega_0)$ and the spatial discretization of the random field $\fbf(\hat{\Xbf})$ are simply referred to as the random field subsequently, and we clarify this difference where necessary.

According to the definition of a Gaussian random field, the joint probability density function (PDF) $p$ for the random field values $\fbf$ at all $\Xbf^{(n_{\rm x})} \in \hat{\Xbf}$  must be a multivariate Gaussian given by
\begin{equation}\label{eq:def-discrete-gauss-random-field}
    \p{\fbf}{\hat{\Xbf}} = {\mathcal{N}}(0; \mathbf{K}) 
    \qquad \mbox{with} \qquad 
    [\mathbf{K}]_{uv} = \mathrm{k}(\Xbf^{(u)}, \Xbf^{\prime(v)}),
\end{equation}
where ${\mathcal{N}}$ represents the multivariate normal distribution with the mean $\mu=0$ and the covariance matrix $\mathbf{K}$ with the components $u$ and $v$ of size $N_{\rm x}$ given by the number of nodes. The components of $\mathbf{K}$ are determined by the chosen covariance function $\mathrm{k}$. We now draw samples from this distribution. In other words, we generate realizations of random fields that satisfy the statistics defined by Eq.~(\ref{eq:def-discrete-gauss-random-field}).

\subsection{Sampling Gaussian random fields} \label{sec:sampling-gaussian-fields}
We now draw samples $\fbf^{(j)}$ from the distribution (\ref{eq:def-discrete-gauss-random-field}). Note the difference between the random field $\fbf$ and samples $\fbf^{(j)}$ of the random field, also known as realizations, generated from random field's PDF (\ref{eq:def-discrete-gauss-random-field}).
In principle, a sample $\fbf^{(j)}$ can be obtained from (\ref{eq:def-discrete-gauss-random-field}) by a Cholesky decomposition of the covariance matrix \cite{Rue2001}, i.e.
\begin{equation}\label{eq:cholesky-sampling}
    \fbf^{(j)} = \mathbf{L} \mathbf{z} \quad \mbox{and} \quad \mathbf{K} 
    = \mathbf{L} \mathbf{L}^{\rm T},
\end{equation}
where $\mathbf{L}$ is the Cholesky factorization of the covariance matrix $\mathbf{K}$, while $\mathbf{z}$ is a vector with dimension equal to the number of nodes $N_{\rm x}$. The components of $\mathbf{z}$ are independent, identically distributed random numbers $\mathrm{z}_{n_{\rm x}}$ from the univariate standard normal distribution, i.e.
\begin{equation}
    \prior{\zbf} = \prod_{{n_{\rm x}}=1}^{N_{\rm x}} \prior{z_{n_{\rm x}}} \quad \mbox{with} \quad 
    \prior{z_{n_{\rm x}}} = {\mathcal{N}}(0;1).
\end{equation}
However, the computational effort of the Cholesky decomposition scales with ${\cal{O}}(N_{\rm x}^3)$ \cite{Kramer2007a} and can therefore be impractical for large or complex domains that require fine discretization, which would lead to a large number of nodes $N_{\rm x}$ and consequently to a large computational effort. 
In general, there are a number of methods for generating realizations of Gaussian random fields \cite{Kramer2007a,Liu2019a}, e.g., (i) spectral methods such as Fourier or Karhunen-Lo\`{e}ve expansions \cite{Shinozuka1991a,Shinozuka1996a,Kim2015a}, (ii) methods based on the solution of a corresponding stochastic (partial) differential equation \cite{Lindgren2011a, Fuglstad2015b, Staber2018a}, or (iii) methods based on iterative procedures and polynomials \cite{Aune2013a,Chow2014a}. Although the latter two methods are more general and can also be applied to more complex geometries, domain discretizations, or non-stationary covariance functions, they are also usually slower. Here we exploit the properties of the simple domain, the discretization (regular grid) and the covariance function (stationarity) to apply the comparatively fast spectral representation method as proposed in \citep{Shinozuka1991a,Shinozuka1996a}. Furthermore, we neglect spatial fluctuations in the circumferential direction of the aorta in order to adopt the procedure for only two spatial dimensions.

For the spectral representation method, we need the Fourier transform of the covariance function, known as the power spectral density ${\cal{S}}$. Using the stationarity of the covariance function, i.e. $\mathrm{k}(\Xbf, \Xbf^\prime) = \mathrm{k}(\Xbf - \Xbf^\prime)$, ${\mathcal{S}}$ can then be calculated using the Wiener--Khinchin theorem \cite{Shinozuka1996a} as
\begin{equation}\label{WienerKhintshineACF1D}
  {\mathcal{S}}(\omegabf) = \frac{1}{(2\pi)^2} \int \limits_{-\infty}^{\infty} 
  \mathrm{k}(\tilde{\Xbf}) \exp\,(- \mathrm{i} \omegabf^{\rm T} \tilde{\Xbf}) \mathrm{d}V_{\scriptsize \tilde{\Xbf}}
  = \varsigma^2 \frac{\iota}{4\pi} \exp \Big(- \frac{\iota^2 \omegabf^2}{4}  \Big),
\end{equation}
where $\tilde{\Xbf} = \Xbf - \Xbf^\prime$ and $\omegabf = (\omega_1, \omega_2)^{\rm T}$.
We may then generate samples $\fbf^{(j)}$ as follows 
\begin{eqnarray} \label{eq:rf-simulation-formula}
    \fbf^{(j)} (\Xbf) & = & \,\sqrt{2} \sum_{n_1=0}^{N_1-1} \sum_{n_2=0}^{N_2-1} A_{n_1,n_2} \Big[ \cos{\Big( n_1 \Delta \omega_1 X_1 +  n_2 \Delta \omega_2 X_2+ \Lambda^{(1)}_{n_1 n_2}\Big)} \nonumber \\
    & & +\cos{\Big( n_1 \Delta \omega_1 X_1 -  n_2 \Delta \omega_2 X_2+ \Lambda^{(2)}_{n_1 n_2}\Big)}\Big],
\end{eqnarray}
with
\begin{equation}
    A_{n_1,n_2} = \sqrt{2 {\cal{S}}(n_1 \Delta\omega_{1}, n_2 \Delta \omega_{2}) \Delta \omega_1 \Delta \omega_2}   
    \qquad \mbox{and} \qquad 
    \Delta \omega_i = \frac{\omega_{{\rm max},i}}{N_i},
\end{equation}
where $X_1$ and $X_2$ denote the coordinates, while $\omega_{{\rm max},i}$ is a cut-off frequency above which ${\mathcal{S}}$ is assumed to be approximately zero. The cut-off frequency $\omega_{{\rm max},i}$ together with $N_i$ defines the discretization of $\omega_i$ to $n_i = 1,\ldots,N_i$ pivot points in equidistant steps $\Delta \omega_i$.
Finally, $\Lambda_{n_1 n_2}^{(i)}$ are the random phase angles, independently uniformly distributed in the interval $[0,2\pi)$. Drawing samples from Eq.~(\ref{eq:def:gp}), i.e.  generating random field realizations, then amounts to choosing a discretization of the Fourier domain, drawing uniform random phase angles and evaluating (\ref{eq:rf-simulation-formula}). Note that to compute a sample that describes a set of values at all locations $\fbf^{(j)} (\hat{\Xbf})$, one has to compute $\fbf^{(j)} (\Xbf)$ in (\ref{eq:rf-simulation-formula}) for all coordinates $\Xbf \in \hat{\Xbf}$ while all other values are fixed, i.e. for fixed random phase angles $\Lambda_{n_1 n_2}^{(i)}$ and fixed Fourier domain discretization. In addition, the sampling algorithm only has to calculate the kernel spectrum $\mathcal{S}$ once.
At this point, we do not restrict ourselves to an equidistant discretization of $\omegabf$, because Eq.~(\ref{eq:rf-simulation-formula}) is the approximation of an integral with a discrete sum, more precisely a Riemann sum.
If one chooses an equidistant discretization of space and Fourier domain, the evaluation of (\ref{eq:rf-simulation-formula}) can be significantly sped up via fast Fourier transforms from ${\cal{O}}(N^2)$ to ${\mathcal{O}}(N\log N)$, where $N=N_1 N_2$, see, e.g., \cite{Shinozuka1991a,Vio2002a, Kramer2007a}. According to \cite{Abrahamsen2018}, samples can then be computed by 
\begin{equation} \label{eq:fft-sampling}
    \fbf^{(j)} = {\cal{Y}}^{-1}\big({\mathcal{S}}^{\frac{1}{2}} {{\mathcal{Y}}(\zbf)}\big),
\end{equation}
where ${\mathcal{Y}}$ is the fast Fourier transform and $\zbf$ is a vector of samples from independent standard normal distributions, analogous to (\ref{eq:cholesky-sampling}).

\subsection{Transforming Gaussian random fields to beta random fields}\label{sec:sampling-uniform-field}
A Gaussian random field $\mathcal{F}$ can be transformed into a non-Gaussian random field $\mathcal{R}$. Let ${\mathcal{C}}_\mathcal{F}$ be the cumulative distribution function of $\mathcal{F}$ and ${\mathcal{C}}_\mathcal{R}^{-1}$ the inverse of the cumulative distribution function of $\mathcal{R}$. Then $\mathcal{R}$ is obtained from the transformation \cite{Grigoriu1995a,Grigoriu1998a, Kim2015a}
\begin{equation}
    \mathcal{R}(\Omega_0) = {\mathcal{C}}^{-1}_{\mathcal{R}}\Big[ {\mathcal{C}}_{\mathcal{F}} \big[\mathcal{F}(\Omega_0) \big]
    \Big]. \notag
\end{equation}
In many cases this equation is difficult to solve because ${\mathcal{C}}_\mathcal{F}$ can be expensive to compute and ${\mathcal{C}}_\mathcal{R}^{-1}$ is often difficult to find.
%
For the Gaussian random fields and the beta random field in this study, an analytical solution is detailed \cite{Vio2001b,Demetriu2005a}. 
Numerical approximations might be required for other types of random field models \cite{Bocchini2008a,Shields2011a,Kim2015a}.

In order to build intuition, we first consider a simple univariate case. If two independent random numbers $\mathrm{f}_1$ and $\mathrm{f}_2$ are distributed according to a Gaussian distribution, then the sum of the square of these numbers, $\mathrm{g} = \mathrm{f}_1^2+\mathrm{f}_2^2$, follows a gamma distribution, more precisely a chi-squared distribution ($\chi^2$-distribution) with two degrees of freedom as a special case of the gamma distribution. Given two independent random variables, $\mathrm{g}_1$ and $\mathrm{g}_2$, that each follow a gamma distribution, then the combination $\beta = \mathrm{g}_1/(\mathrm{g}_1 + \mathrm{g}_2)$ follows a beta distribution, see \ref{app:proof-beta-margin}. 
After that, the hyperparameters of a beta distribution can be chosen in such a way that we get a uniform distribution. We now use an analogous result generalized to random fields \cite{Hasofer1998a,Demetriu2005a,Vio2001b,Vio2002a}, in order to construct a uniform random field from Gaussian random fields.

Let $\{\fbf_{r}(\hat{\Xbf})\}$, $r = 1,\dots,2s$, $s \in \mathbbm{N}$, be a collection of independent Gaussian random fields. However, we choose here that the random fields $r$ are identical yet still independent. For the sake of simplicity, we omit the superscripts $j$ in the following, which indicate samples of the random field $r$, as used analogously in Section~\ref{sec:sampling-gaussian-fields}.
Then, gamma random fields $\gbf_{s}(\hat{\Xbf})$ are computed as
\begin{equation}
    \label{eq:Gammafield_definition}
     \gbf_{s}(\hat{\Xbf}) = \frac{1}{2} \sum_{r=1}^{2s} \fbf_{r}^2(\hat{\Xbf}).
\end{equation}
Based on this, one is able to generate a gamma field sample from one sample each of at least two independent Gaussian random fields, $\fbf_{1}(\hat{\Xbf})$ and $\fbf_{2}(\hat{\Xbf})$.
The correlation structure for this gamma field $\mathrm{k}_{g_s}(\tilde{\Xbf})$ is transformed as follows
\begin{align}
    \mathrm{k}_{g_s}(\tilde{\Xbf}) = \big(\mathrm{k}(\tilde{\Xbf})\big)^2,
\end{align}
where $\mathrm{k}(\tilde{\Xbf})$ is the stationary covariance function for the Gaussian field used in Eq.~(\ref{eq:rf-simulation-formula}).
Note that the gamma fields also contain exponential or $\chi^2$-distributions as a special case. 
With a set of samples from two independent gamma fields, $\gbf_{s}(\hat{\Xbf})$ and $\gbf_{s^\prime}(\hat{\Xbf})$, 
each characterized by the same covariance function, it is possible to sample a beta random field $\betabf_{s,{s^\prime}}(\hat{\Xbf})$
from
\begin{equation}\label{eq:Beta_random_field_definition}
    \betabf_{s,{s^\prime}}(\hat{\Xbf}) = \frac{\gbf_s(\hat{\Xbf})}{\gbf_s(\hat{\Xbf}) + \gbf_{s^\prime}(\hat{\Xbf})}.
\end{equation}
The univariate marginal PDF of (\ref{eq:Beta_random_field_definition}) is a beta distribution, see \ref{app:proof-beta-margin}, which is given by 
\begin{equation} \label{eq:beta-marginal}
    \p{\beta}{\Xbf} = \frac{1}{{\cal{B}}(s,s^\prime)} \beta^{s-1} (1-\beta)^{{s^\prime}-1}, 
    \qquad 0 \leq \beta \leq 1,
\end{equation}
where ${\mathcal{B}}$ is the beta function, while $\beta(\Xbf)$ is a univariate random variable at a single particular location $\Xbf$. This means that the random variable $\beta$ is the value of the random field $\betabf(\hat{\Xbf})$ at location $\Xbf$ and that $\beta$ follows a beta distribution at all locations $\Xbf$.
The correlation structure of this beta random field, $\mathrm{k}_{\beta_{s,s^\prime}}$, is
\begin{align}
    \mathrm{k}_{\beta_{s,s^\prime}}(\tilde{\Xbf}) &= 1 - (s+s^\prime) \bigg( \frac{1-\mathrm{k}(\tilde{\Xbf})}{-\mathrm{k}(\tilde{\Xbf})}\bigg)^{s+s^\prime} \bigg[\log \big(1-\mathrm{k}(\tilde{\Xbf})\big) - \sum_{l=1}^{s+s^\prime-1}\frac{1}{l}\bigg(\frac{-\mathrm{k}(\tilde{\Xbf})}{1-\mathrm{k}(\tilde{\Xbf})} \bigg)^l \bigg],
\end{align}
where $s+s^\prime >1$ and again $\mathrm{k}(\tilde{\Xbf})$ is the stationary covariance function for the Gaussian field used in Eq.~(\ref{eq:rf-simulation-formula}).
For the special case $s = s^\prime = 1$ we get a uniform distribution in the interval $\beta \in [0,1]$, i.e.
\begin{equation}
    \p{\beta}{\Xbf} = \mbox{const.}
\end{equation}
This means that by this procedure we get a non-Gaussian random field for which every marginal PDF is a uniform distribution.
The degradation parameter $\xibf$, as defined in \eqref{eq:degradation-parameter}, is bounded, a property correctly modeled by a beta random field. Due to the limited experimental data available, a uniform random field was chosen. With Eq.~(\ref{eq:Beta_random_field_definition}) we can now define the degradation parameter as follows
\begin{equation} \label{eq:def-xi-field}
    \xibf(\hat{\Xbf}) := \betabf_{1,1}(\hat{\Xbf}).
\end{equation}
We therefore found a way to draw samples from the desired probability distribution
\begin{equation}
    \p{\xibf}{\hat{\Xbf}}.
\end{equation}
As summarized in Algorithm~\ref{algo1}, this is achieved by drawing samples of the auxiliary Gaussian random field (\ref{eq:def:gp}) using the covariance function defined in (\ref{eq:def-kernel}) via the sampling scheme (\ref{eq:rf-simulation-formula}) or (\ref{eq:fft-sampling}), respectively. The Gaussian random field samples are then input into (\ref{eq:Gammafield_definition}), yielding samples of a gamma-type random field. Finally, samples of the gamma-type random field are used in (\ref{eq:Beta_random_field_definition}), resulting in samples of a beta random field.
Here, with $s = s^\prime = 1$, we need two Gaussian random field samples to calculate a gamma-type random field sample. Then we need two such gamma-type random field samples to compute a beta random field $\xibf$. 
In the special case of the uniform field in (\ref{eq:def-xi-field}), $s = s^\prime = 1$, strictly speaking, a total of four Gaussian random field samples are required to calculate one beta random field sample.

Note that non-uniform beta fields can be modeled for general bounded parameters by choosing $s$ and $s^\prime$ accordingly. In addition, the gamma field mapping (\ref{eq:Gammafield_definition}) and the beta field mapping (\ref{eq:Beta_random_field_definition}) do not allow negative correlations. This limitation is not inherent to the respective fields, but to the special mappings shown in (\ref{eq:Gammafield_definition}) and (\ref{eq:Beta_random_field_definition}).
\begin{algorithm}[t]
\caption{
This pseudo-code generates $j = 1,\ldots,N_j$ samples $\betabf_{s,s^\prime}^{(j)}$ of beta random fields, where the sample index $j$ as in Section~\ref{sec:sampling-gaussian-fields}, is re-introduced here. Anticipating the quantities introduced in Section~\ref{s:3}, we refer to $N_j = N_{\mathrm{s}}$ in the context of the Monte Carlo integration and $N_j = N_{\xi}$ in regard to NN training data, see, e.g., Eqs.~(\ref{eq:stochastic-integration-brute-force}), (\ref{eq:final-result}) and (\ref{eq:def-data}).
%
} \label{algo1}
\begin{algorithmic}[1]
\STATE BETA\_RANDOM\_FIELD\_SAMPLES ($N_j,s,s^\prime, \hat{\Xbf}$):
    \FOR{$j = 1,\ldots,N_j$}
        \FOR{$r = 1,\ldots,2s$}
            \STATE{Generate $\fbf_r^{(j)}$}  \COMMENT{see (\ref{eq:rf-simulation-formula}) or (\ref{eq:fft-sampling})} 
        \ENDFOR
        \STATE{$\gbf_s^{(j)} \gets \frac{1}{2} \sum_{r=1}^{2s} \big( \fbf_{r}^{(j)}\big)^2 $ \COMMENT{see (\ref{eq:Gammafield_definition})}}
        \FOR{$r^\prime = 1,\ldots,2s^\prime$}
            \STATE{Generate $\fbf_{r^\prime}^{(j)}$ \COMMENT{see (\ref{eq:rf-simulation-formula}) or (\ref{eq:fft-sampling})}} 
        \ENDFOR
        \STATE{$\gbf_{s^\prime}^{(j)} \gets \frac{1}{2} \sum_{r=1}^{2s^\prime} \big(\fbf_{r^\prime}^{(j)}\big)^2  $ \COMMENT{see (\ref{eq:Gammafield_definition})} }
        \STATE{$\betabf_{s,{s^\prime}}^{(j)} \gets \frac{\gbf_s^{(j)}}{\gbf_s^{(j)}+\gbf_{s^\prime}^{(j)}}$ \COMMENT{see (\ref{eq:Beta_random_field_definition})}}
        \STATE{$\xibf^{(j)} \gets \betabf_{s,{s^\prime}}^{(j)}$ }
        \STATE{Compute Cauchy stresses $\Sigma^{(j)}$ for given $\xibf^{(j)}$ with FE  or NN}  
    \ENDFOR
\end{algorithmic}
\end{algorithm}

\subsection{Application to a boundary-value problem}\label{sec:boundary-value-problem}
\begin{figure}[t!]
	\centering
	\includegraphics[width=\linewidth]{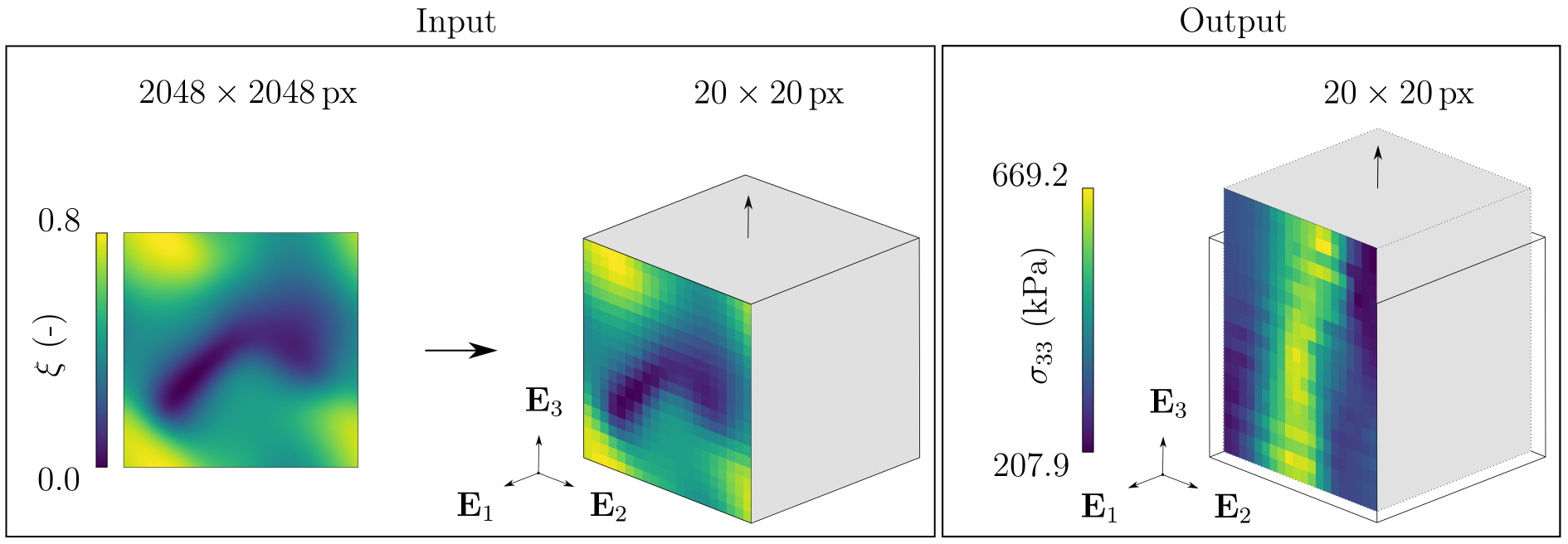}
	\vspace*{-0.75cm}
	\caption{
	Application of a representative random field of the degradation parameter $\xi$ to a uniaxial extension test of a unit cube in connection with the material properties of the aortic wall \cite{RolfPissarczyk2021a}. The unit cube, represented in the reference configuration (input) and the intermediate configuration (output), is aligned with the Cartesian unit basis vectors $\Ebf_1, \Ebf_2$ and $\Ebf_3$ and subjected to a displacement load in the radial direction $\mathbf{E}_3$. The Cartesian unit basis vectors $\Ebf_1$ and $\Ebf_2$ correspond to the circumferential and axial directions of the aortic wall, respectively. The input random field is mapped from a high resolution field ($2048\times2048$\,px) to a low resolution field ($20\times20$\,px) such that the evaluation points of the random field coincide with the Gaussian integration points of the FEs. The output of the FE analysis then also provides the Cauchy stress component $\sigma_{33}$ in a low resolution field ($20\times20$\,px).
	}
	\label{fig:randomfield}
	\vspace*{-0.2cm}
\end{figure}

%
The stochastic constitutive model is applied to a boundary-value problem, which is then solved using the FE method. More precisely, in the FE analysis program FEAP~\cite{FEAP}, a uniaxial extension test of an incompressible unit cube defined by dimensions $1\times1\times 1$\,mm$^3$ 
is performed as shown in Fig.~\ref{fig:randomfield}. The unit cube is aligned with the Cartesian unit basis vectors $\Ebf_1, \Ebf_2$ and $\Ebf_3$ and a uniform displacement of $0.4$\,mm along the top face is applied so that the loading direction coincides with the radial vector, i.e. $\Ebf_3 = \Ebf_{\rm R}$. Here we define $\Ebf_1$ and $\Ebf_2$ as circumferential and axial directions, respectively. We discretized the unit cube with $100$ 8-node hexahedral mixed Q1/P0 elements, ten elements in the $\Ebf_2$ and $\Ebf_3$ directions, respectively, and one element in the $\Ebf_1$-direction. 

The augmented Lagrangian method in FEAP~\cite{FEAP} was applied to ensure incompressibility. To model the inherent microstructure of the aortic wall, two families of collagen fibers, an isotropic ground substance and one family of elastic fibers were defined. The mean fiber directions of the two collagen fiber families are defined in the ($\Ebf_2$-$\Ebf_3$) plane with a symmetric in-plane angle around $\Ebf_2$. In contrast, the mean fiber direction of the family of elastic fibers is aligned with the $\Ebf_3$-direction, so the dispersion of elastic fibers realistically reflects both the elastic lamellar and the interlamellar elastic fibers. The local degradation of elastic fibers is then taken into account by changing the degradation parameter, resulting in a locally reduced delamination strength. The mechanical and structural parameters of the respective constituents agree with the computational study by Rolf-Pissarczyk et al.~\cite{RolfPissarczyk2021a}.

The input for the FE analysis are uniform random fields describing the local elastic fiber degradation generated by the spectral method as described in Section~\ref{sec:sampling-gaussian-fields} and the subsequent transformation described in Section~\ref{sec:sampling-uniform-field}. The correlation length of the degradation parameter was chosen to be $\iota = \sqrt{2}/3$\,mm and the simulated noise added in the random field sample was chosen to be $\varsigma^2 = 0.173$. To simplify the computational problem and save computational costs, we assume that the degradation parameter varies only in the ($\Ebf_2$-$\Ebf_3$) plane. 
Therefore, a two-dimensional random field was simulated, then duplicated, and finally two layers were stacked back-to-back for the three-dimensional unit cube. A three-dimensional FE simulation was then carried out with a two-dimensional random field, as shown in Fig.~\ref{fig:randomfield}. In addition, the two-dimensional random field was sampled on an equidistant grid of size $2048\times 2048$\,px and further sampled down to a size of $20\times 20$\,px, so that the evaluated grid points of the low-resolution image coincide with the non-equidistant Gaussian integration points of the FE analysis. More specifically, a Gaussian quadrature rule of second order was applied. The random field values can usually also be generated directly for the integration points, but the grid of the integration points is usually not uniform.

Figure~\ref{fig:randomfield} shows a representative example of the Cauchy stress component $\sigma_{33}$. To compare the input and output at the same locations, we evaluated the Cauchy stress only at the Gaussian integration points instead of mapping the stress onto the nodes as usual. Since the remaining Cauchy stress components are small compared to $\sigma_{33}$, we have neglected them in this study. Also note that in some cases the FE analysis did not converge, which was usually observed when the gradient between adjacent degradation parameters of the random field was particularly high.

\section{Uncertainty propagation to stress distributions}\label{s:3}
The stochastic model entails an uncertainty in the stress distribution resulting from a uniaxial extension test. A comprehensive overview of the methods for quantifying these uncertainties in the stress distribution, i.e. on the propagation of the uncertainties through the model, can be found in \cite{Ghanem2017a}. Here, we choose a Bayesian approach \cite{vonderLinden2014a}. 
For readers unfamiliar with Bayesian probability theory, a brief introduction to the basic rules of probability theory is given and then applied to formulate the UQ problem. Subsequently, NNs are introduced as surrogate models, specifically the Bayesian encoder-decoder architecture to solve the UQ problem. We also refer to the literature \cite{Jaynes2003a,Sivia2006a,VonToussaint2011a,vonderLinden2014a}.

\subsection{Probability theory and uncertainty quantification} \label{sec:bayesics}
Let $Q$, $R$ and $P$ be three propositions. If, e.g., $Q$ is a continuous random variable, such a proposition could be \lq $Q$ has a certain value'. We will also need other types of propositions, such as the proposition that a certain model $P$ is true or a certain new data set $R$ was measured. We can further combine propositions with Boolean algebra into new propositions, e.g., $Q$ AND $R$ is the proposition that \lq $Q$ has a certain value AND $R$ has been measured'. 
In the following, such AND-compositions are denoted with a comma, e.g., $(Q,R)$. Propositions can also be conditioned on each other, e.g., $Q\mid R$ reads \lq $Q$ has a certain value given $R$ has been measured'. Hence, $Q \mid R, P$ reads \lq $Q$ has a certain a certain value given $R$ has been measured AND model $P$ is true'.
The basic rules hold for all these types of propositions. The first rule is Bayes' theorem
\begin{equation}\label{eq:bayes-theorem}
    \p{Q}{R, P} = \frac{\p{R}{Q, P}\p{Q}{P}}{\p{R}{P}},
\end{equation}
where $\p{Q}{R,P}$ is the conditional probability for $Q$ given that $R$ AND $P$ is known. 
The probability $\p{Q}{R,P}$ is usually called {\em posterior}. 
In this study, $Q$ are the mechanical stresses whose uncertainties we want to quantify, $R$ is a data set, and $P$ are the model assumptions.
The probability $\p{R}{Q,P}$ is called the {\em likelihood}, e.g., the probability for measuring a data set $R$ given that the model $P$ is true and $Q$ is the true value. $\p{Q}{P}$ is the {\em a priori} probability for $Q$, i.e. the probability before the new data measurements were taken into account, and $\p{R}{P}$ is the evidence, which is a normalization constant. The second basic rule is the marginalization rule. For continuous variables $Q$ it reads
\begin{equation} \label{eq:marginalization_rule}
    \p{R}{P} = \int \p{R}{P,Q}\p{Q}{P} \mathrm{d}V_Q,
\end{equation}
which is to be understood as a volume integral over the domain of $Q$.

According to Section~\ref{s:2}, the Cauchy stress tensor is a function of the random field parameter (the collection $\xibf$ of degradation parameters) and the location $\Xbf$, i.e. $\sigbf (\xibf, \Xbf)$. In the following, we consider a particular component $\sigma_{\nu \nu^\prime}$ of the Cauchy stress tensor. 
To simplify the notation and because $\nu,\nu^\prime$ can be chosen arbitrarily, we now suppress the indices for the tensor component and write $\obs \equiv \obs_{\nu\nu^\prime}$. It is possible to extend the approach to consider all stress tensor components together. Note that the quantity of interest, here $\obs$, can in general be any quantity of interest derived from the FE analysis, such as the displacement field.
The uncertainty of a quantity of interest $\obs$, here a particular component of the Cauchy stress tensor, at a fixed current location $\Xbf$,
given the degradation field $\xibf(\hat{\Xbf})$ at all reference locations $\hat{\Xbf}$, is described by the probability density function for the value of $\obs$, i.e.
\begin{equation}\label{eq:starting-point}
    \p{\obs}{\xibf, \Xbf},
\end{equation}
where we have used $\p{\obs}{\xibf, \hat{\Xbf}}= \p{\obs}{\xibf, \Xbf}$, i.e. given the full random field realization $\xibf$ at all $\Xbf \in \hat{\Xbf}$, one only has to specify the measurement location $\Xbf$.

However, according to Section~\ref{s:2}, the degradation parameter $\xibf$ is a random field and therefore not exactly known.
According to the Bayesian paradigm, we have to average over all unknowns, meaning to marginalize the unknowns as in (\ref{eq:marginalization_rule}). Using the PDF for the random field $\xibf$ as described in Section~\ref{s:2} this is achieved via
\begin{eqnarray} 
    \p{\obs}{\Xbf} & = & \int \p{\obs}{\xibf, \Xbf} \p{\xibf}{\hat{\Xbf}}  \mathrm{d}V_{\xi} \label{eq:def-sought-for-pdf} \\ 
    & & \approx \sum_{n_{\rm s}=1}^{N_{\mathrm{s}}} \delta\Big(\obs -  \obs(\xibf ^{(n_{\mathrm{s}})}, \Xbf) \Big) W_{\xi^{(n_{\rm s})}}, \label{eq:stochastic-integration-brute-force}
\end{eqnarray}
where (\ref{eq:stochastic-integration-brute-force}) is an approximation of (\ref{eq:def-sought-for-pdf}), obtained by stochastic integration. In other words, a Monte Carlo integration \cite{Kroese2011a} was applied with $N_{\mathrm{s}}$ samples, often denoted as particles, with sample weights, specifically the probability mass $W$ and the Dirac delta function $\delta$. 
Note that each sample $\xibf^{(n_{\rm s})}$ here corresponds to a random field realization as introduced before.
Here we took advantage of the fact that we can draw samples from the distribution $\p{\xibf}{\hat{\Xbf}}$, as shown in Sections~\ref{sec:sampling-gaussian-fields} and \ref{sec:sampling-uniform-field}.
Stochastic integration then amounts to generating $N_{\mathrm{s}}$ random field realizations $\xibf^{(n_{\rm s})}$, computing the corresponding stress component each $\obs^{(n_{\rm s})}:=\obs(\xibf^{(n_{\rm s})}, \Xbf)$, and aggregating the results. For (\ref{eq:stochastic-integration-brute-force}), we later calculate $\obs^{(n_{\rm s})}$ with a trained NN instead of with the expensive FE method.
In the simplest form of Monte Carlo, the weights are all equal, i.e. $W_{\xi^{(n_{\rm s})}} = 1/N_{\mathrm{s}}$. Note that $W$ are sample weights in the sense of the relative probability mass and are not to be confused with the weights in the sense of surrogate parameters, which define the NN to be introduced later. An important advantage of stochastic integration over deterministic integration is that its convergence rate is independent of the integral dimension $N_{\rm x}$.

After determining (\ref{eq:def-sought-for-pdf}) one can also compute the mean, standard deviation, variance and covariance, where the expectation value is denoted as $\expect{\cdot}$. Therewith,
\begin{eqnarray}
    \expect{\obs(\Xbf)} & = & \frac{1}{{\mathcal{Z}}}\int \obs \; \p{\obs}{\Xbf} \mathrm{d}{\obs}, \\
    \mathrm{std}\,(\obs(\Xbf)) & = & \sqrt{\mathrm{var}\,(\obs(\Xbf))}  \label{eq:def-std},\\
    \mathrm{var}\,(\obs(\Xbf)) & = & \frac{1}{{\mathcal{Z}}}\int [\obs - \expect{\obs(\Xbf)}]^2 \p{\obs}{\Xbf} \mathrm{d}{\obs}, \\
    \mathrm{cov}\,(\obs(\Xbf), \obs(\Xbf^\prime)) &= & \frac{1}{{\mathcal{Z}}}\int [\obs - \expect{\obs(\Xbf)}]  [\obs - \expect{\obs(\Xbf^\prime)}]   \p{\obs}{\Xbf} \mathrm{d}{\obs},
\end{eqnarray}
with a normalization ${\mathcal{Z}}$. In addition, the probability that the random variable $\obs$ exceeds a critical threshold value $\sigma_{\rm crit}$ at a fixed location $\Xbf$ is
\begin{equation}\label{eq:local-prob-exceed-threshold}
        P(\obs > \sigma_{\rm crit} \mid \Xbf) = \int^{-\infty}_{\sigma_{\rm crit}}\p{\obs}{\Xbf} \mathrm{d}{\obs} =1- \int_{-\infty}^{\sigma_{\rm crit}}\p{\obs}{\Xbf} \mathrm{d}{\obs}.
\end{equation}
If the critical threshold value $\sigma_{\rm crit}$ is the failure stress of the aortic wall, then Eq.~(\ref{eq:local-prob-exceed-threshold}) is the local failure probability at a certain location $\Xbf$. Next, the global failure probability at any location $\Xbf^{(n_{\rm x})} \in \hat{\Xbf}$ reads
\begin{equation}\label{eq:global-prob-exceed-threshold}
     P(\obs > \sigma_{\rm crit}) \approx \frac{1}{N_{\rm x}} \sum_{n_{\rm x}=1}^{N_{\rm x}}{ P(\obs > \sigma_{\rm crit} \mid \Xbf^{(n_{\rm x})}) }.
\end{equation}

Solving the problem given in (\ref{eq:def-sought-for-pdf}) with (\ref{eq:stochastic-integration-brute-force}) leads to two problems.
First, the solution of the FE analysis, i.e. the Cauchy stress component $\obs$, can be computationally rather expensive for a large number of degrees of freedom or multi-physics coupling.
Then the computational effort limits the number of model evaluations $N_{\mathrm{s}}$, which in turn restricts the accuracy of the estimate (\ref{eq:stochastic-integration-brute-force}). The accuracy of the estimator (\ref{eq:stochastic-integration-brute-force}) is proportional to $\sqrt{N_{\mathrm{s}}}$. In other words, if one wants to reduce the variance of the Monte Carlo estimate (\ref{eq:stochastic-integration-brute-force}) by one digit, then ten times as many samples are required.
Second, the computational effort to generate the random field $\xibf$ scales unfavorably with the domain discretization $\hat{\Xbf}$ in the first place. 
While the second problem has already been addressed \cite{Panunzio2018a, DeCarvalhoPaludo2019a}, we instead focus on how the first problem can be addressed by a so-called surrogate model or meta-model. Next, we introduce NNs as surrogate models in Section~\ref{sec:nn-surrogates}. 
Section~\ref{sec:wedding} makes it clear how such a surrogate can be used to solve (\ref{eq:stochastic-integration-brute-force}).

\subsection{Neural networks as a surrogate and the encoder-decoder architecture} \label{sec:nn-surrogates}
CNNs have seen an increase in interest and applications in the fields of computer vision and pattern recognition since the ImageNet challenge~\cite{Deng2009a} and to date most published research still involves the training of a CNN~\cite{Tajbakhsh2020a,Zhang2021a}. This popularity also brought CNNs growing attention in other fields, including shape modeling~\cite{Cao2020a}, computational chemistry~\cite{Goh2017a}, and physics-based simulations~\cite{Mendizabal2020a}.
NNs generally consist of fully connected graphs where the nodes or neurons are interconnected by weighted arcs or synapses. The actual configuration is sought through an iterative process called training. 

Fully connected graphs can be difficult to train on large inputs and may require high computational and memory costs. Convolution can be used to analyze spatial correlations between subsets of inputs, like neighboring cells in a matrix. This resulted in CNNs represented as stacked layers of linear convolution followed by nonlinear activation~\cite{Pepe2021a}. The NN $\mathcal{M}$ is therefore a nested sequence of functions  $\mathcal{M}^{(\ell)}$, i.e.
\begin{equation} \label{eq:network-chain}
    \mathcal{M} = \mathcal{M}^{(L)} \circ \mathcal{M}^{(L-1)} \circ \cdots \circ \mathcal{M}^{(1)} \circ \mathcal{M}^{(0)},
\end{equation}
which yields a recursive relation for the layers $\ell=1,\ldots,L$, with neurons $n_\ell = 1,\ldots,N_\ell$ in each layer according to
\begin{equation} \label{eq:network-recursive}
    \mathcal{M}^{(\ell)} = \mathbf{h}_{\ell, n_\ell} \bigg( \sum_{n_\ell=1}^{N_\ell} \wbf_{(\ell-1,n_\ell)} \odot \mathcal{M}^{(\ell-1)} + \mathbf{b}^{(\ell-1)} \bigg), 
    \qquad \mathcal{M}^{(0)} := \xibf(\hat{\Xbf}),
\end{equation}
where $\mathbf{h}$ is the above-mentioned nonlinear activation function of neuron $n_\ell$ in layer $\ell$ applied element-wise to its matrix- or vector-valued argument, while $\wbf_{(\ell,n_\ell)}$ are the elements of the set of weight matrices $\wbf$, and $\mathbf{b}$ are additional parameters called biases. Note that $\mathcal{M}^{(\ell)}$ is generally matrix-valued since $\xibf$ are also matrix-valued. The weight matrices are parameters for the NN and should not be confused with the particle weights $W_{\xi^{(n_{\rm s})}}$ from Section~\ref{sec:bayesics} or $W_{\mathrm{w}^{(n_{\rm m})}}$ introduced later in Section~\ref{sec:variational-inference}. 
Additionally, $\wbf_{(\ell,n_\ell)}$ is a matrix of weights that is multiplied element-wise, $\odot$, with the output from the previous layer $\mathcal{M}^{(\ell-1)}$. The indices $u$ and $v$ in $[\wbf_{(\ell,n_\ell)}]_{uv}$ then denote the weights for particular features $[\mathcal{M}^{(\ell-1)}]_{uv}$, and summation with respect to $n_\ell$ forms the convolution. For example, $[\mathcal{M}^{(0)}]_{uv}$ here are the values of the random field $\xibf$ at location  $(X_1^{(u)}, X_2^{(v)})^T$. 
Note that the argument of $\mathbf{h}$ represents a generalized discrete convolution with $\prod_\ell n_\ell$ convolution kernel parameters. 
Also, the convolution kernel should not be confused with the Stein kernel to be introduced in Section~\ref{sec:variational-inference}. For fully connected layers, all corresponding weights would be non-zero and independent, i.e. all neural connections are retained, while in a convolutional layer the weights of adjacent neurons $n_\ell, n_{\ell+1}, n_{\ell-1}, n_{\ell+2},\ldots$ are shared, effectively reducing the number of free parameters. 

The training process then boils down to choosing an optimization criterion, e.g., the ${\mathcal{L}}_1$ or ${\mathcal{L}}_2$ loss functions, and minimizing this loss function with respect to all weights $\wbf$ and biases $\mathbf{b}$ with a suitable optimization algorithm. The loss function and optimization implications are discussed in more depth in Section~\ref{sec:wedding}. It can be shown that the gradients for feed-forward networks can be efficiently evaluated. These gradients are then used for error back-propagation and weight update during training~\cite{Goodfellow2016a}. 
This procedure makes it possible to automatically extract multi-scale features from high-dimensional input, reducing the need for hand-crafted feature engineering, such as searching for the right set of basis functions or relying on experts knowledge~\cite{Zeiler2014a}. In our case, these features will primarily be complex spatial correlations of the stress distributions.

We usually speak of deep CNNs when the network has two or more intermediate layers~\cite{Hinton2006a,LeCun2015a}.
Although deep CNNs have shown high accuracy on a large number of tasks, a limitation stems from their low robustness and reproducibility~\cite{HaibeKains2020a,Li2020a}, also given their intrinsic inability to express uncertainty \cite{Zhu2018a}.
A solution to this problem is provided by Bayesian deep networks, which also allow the prediction uncertainty to be expressed \cite{MacKay1992a,Neal1996a,Gal2016a}. 
A Bayesian network can quantify the predictive uncertainty by treating the network parameters as random variables and by performing Bayesian inference on those uncertain parameters, even when the training data set is small. In a fully Bayesian treatment, one would rather learn a probability distribution than minimize a loss function, as will be introduced in Section~\ref{sec:variational-inference}.

%
Depending on the learning task and data structure, a large number of NN architectures have been proposed \cite{Shrestha2019a,Sengupta2020a}. A common architecture among them is the encoder-decoder~\cite{Tajbakhsh2020a,Mendizabal2020a,Pepe2021a}.
An encoder-decoder architecture is typically used when data needs to be compressed and decompressed, as in matrix-to-matrix regression tasks. Image-to-image transformations are common examples~\cite{Pepe2021a}. 
The task of reducing the complexity of the input data, i.e. the selection or extraction of features, is called an encoder, while the reverse process of decompression is called a decoder.
The entire process of reducing the number of features to an encoder space or latent space is understood as dimensional reduction.
An encoder-decoder architecture is considered good if it retains the maximum information when encoding, while showing minimal error when reconstructing the data in the decoder. 

The encoder-decoder architecture has shown a growing number of applications since its introduction \cite{Sutskever2014a,LeCun2015a}.
In this study we use an architecture similar to \cite{Zhu2018a}, which showed state-of-the-art performance in terms of prediction accuracy and in comparison to other established approaches, such as Gaussian processes \cite{Zhu2018a}. 
An important motivation for this choice of a surrogate model is that the encoder-decoder structure allows to extract multi-scale features and spatial correlations from the input. These features are then processed by the decoder and finally preserved in the output.

\begin{figure}[h!]
    \centering
    \includegraphics[width=0.96\textwidth]{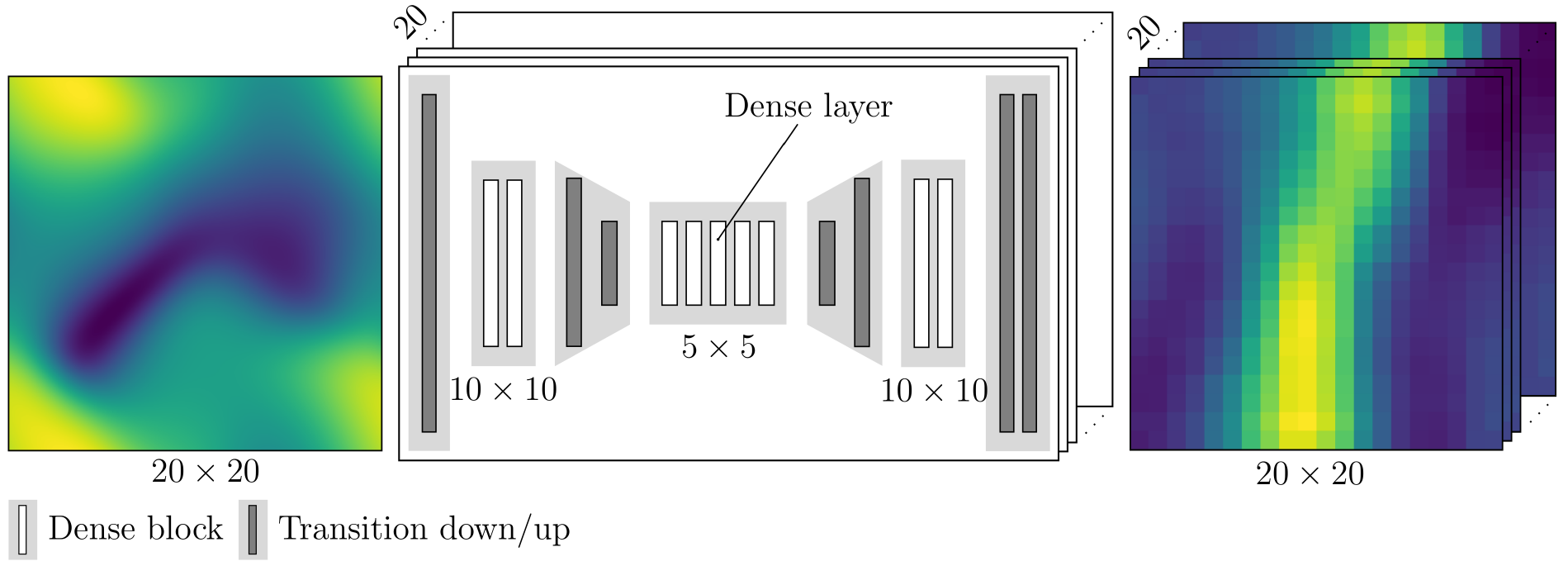}
    \vspace*{-0.3cm}
    \caption{
    The illustration outlines the architecture of the NN. A representative random field on the left is first convoluted to enter the encoder. Then, after entering the dense block with two dense layers, it is transitioned down and passes through a dense block with five dense layers. Subsequently, the representative random field is transitioned up again, followed by a dense block with two dense layers, the decoder, and a final convolution. The final output on the right is the network's prediction for the spatial distribution of the Cauchy stress component $\sigma_{33}$. To predict a single random field, we use an ensemble of 20 NNs connected in parallel, from which then follows that we also obtain 20 NN predictions.}
    \label{fig:architecture-overview}
    \vspace*{-0.35cm}
\end{figure}

Figure~\ref{fig:architecture-overview} shows the encoding path that takes random field realizations and passes them through a convolution layer. 
The extracted feature maps are passed to a series of dense blocks and encoding layers.
After the last dense block and a transition layer, the high-level coarse feature maps are passed through dense layers and then fed into the decoding path of Fig.~\ref{fig:architecture-overview}. 
The decoding path has a similar structure to the encoder, but with decoding layers instead.
At the end of the last decoding layer, predictions of the $\obs$ output fields are made.

For more details on network architecture, network layers and layer parameters, we refer the reader to \ref{app:details-architecture}.

\subsection{Bayesian uncertainty propagation with a neural network surrogate model} \label{sec:wedding}
Next, we describe how we can solve the uncertainty propagation problem formulated in Eq.~(\ref{eq:stochastic-integration-brute-force}) with the help of a surrogate model, here an encoder-decoder CNN, as introduced in Section~\ref{sec:nn-surrogates}, with a Bayesian approach. 
It is assumed that the results of the FE analysis, here the Cauchy stress component $\obs$, can be approximated by a parametrized function, i.e. the surrogate model $\mathcal{M}$
\begin{equation} \label{eq:def-surrogate}
  \obs \approx \mathcal{M}(\xibf ; \wbf \mid \Xbf),
\end{equation}
where $\wbf$ is a set of surrogate parameters. Also, we will use the  encoder-decoder CNNs introduced above as surrogate models, i.e. $\mathcal{M}$ has the form of Eq.~(\ref{eq:network-chain}).
In connection with NNs, $\wbf$ are often called weights, which should not be confused with the stochastic integration weights $W$ mentioned above. 
Then the inputs and outputs of the FE analysis are collected in a training data set ${\cal{D}}$ according to
\begin{equation}
    {\mathcal{D}} = \{\hat{\Xbf}, \bm{\Xi}, \bm{\Sigma}\},
\end{equation}
which is grouped as follows
\begin{equation}\label{eq:def-data}
    \hat{\Xbf} = \{\Xbf^{(n_{\rm x})}\}_{n_{\rm x}=1}^{N_{\rm x}}, \quad \bm{\Xi} = \{\bm{\xi}^{(n_{\xi})}\}_{n_{\xi}=1}^{N_{\xi}}, \quad
    \bm{\Sigma} = \{\obssmallset\}_{n_{\xi}=1}^{N_{\xi}}, \quad \obssmallset = \{\obs^{(n_{\rm x},n_{\xi})}\}_{n_{\rm x}=1}^{N_{\rm x}},
\end{equation}
where ${\mathcal{D}}$ consists of $N_{\xi}$ random field samples and their corresponding stresses computed with the FE method at all $N_{\rm x}$ measurement points or nodes in the domain, $\obs^{(n_{\rm x},n_{\xi})}:=\obs(\xibf^{(n_{\xi})}, \Xbf^{(n_{\rm x})})$.
Note that we distinguish between the number of random field realizations $N_{\xi}$ for which a corresponding stress distribution is computed using the FE method and then used to create a surrogate model later, and the number of Monte Carlo samples $N_{\mathrm{s}}$ for the stochastic integration in Eq.~(\ref{eq:stochastic-integration-brute-force}). 
Accordingly, we also distinguish the indices $n_{\xi} = 1,\ldots,N_{\xi}$ and $n_{\mathrm{s}}=1,\ldots,N_{\mathrm{s}}$. The $N_{\xi}$ random field samples $\xibf^{(n_{\xi})}$ together with the corresponding stress distributions obtained with the FE method $\obssmallset$ will shortly be used to learn a surrogate model. For the $N_{\mathrm{s}}$ random field samples $\xibf^{(n_{\rm s})}$ in the stochastic integration of (\ref{eq:stochastic-integration-brute-force}), we will compute the stress distributions with the surrogate model learned in this way. 

Next, an optimality criterion ${\mathcal{O}}$ is defined to determine the \lq optimal' surrogate parameters $\wbf^{\ast}$ based on the given training data ${\mathcal{D}}$, i.e.
\begin{equation} \label{eq:optimization-problem}
    \wbf^{\ast} = \arg \max_{\wbf} {\mathcal{O}}\Big( \mathcal{M}(\xibf; \wbf \mid \Xbf); {\mathcal{D}}\Big).
\end{equation}
This optimization process corresponds to the NN training mentioned in Section \ref{sec:nn-surrogates}.
The optimal surrogate parameters could, {\em inter alia}, be found by minimizing the sum of the least squares, i.e. {$\wbf^{\ast}=\arg \min_{\wbf} \sum_{n_{\rm x}=1}^{N_{\rm x}} \sum_{n_{\xi}=1}^{N_{\xi}} \Big(\obs^{(n_{\rm x},n_{\xi})} - \mathcal{M}(\xibf^{(n_{\xi})}; \wbf \mid \Xbf^{(n_{\rm x})})\Big)^2$}. 
As soon as $\wbf^{\ast}$ and thus $\mathcal{M}$ are determined, one then substitutes the surrogate model for the FE analysis.
The surrogate model can then make predictions for the FE results with significantly reduced computational costs. 
For the intended purpose, this means that we substitute the PDF for a particular component of the Cauchy stress tensor (\ref{eq:starting-point}) with a surrogate-based PDF, i.e.
\begin{equation}
    \p{\obs}{\xibf, \Xbf} \to \p{\obs}{\xibf, \Xbf, {\mathcal{D}}, \mathcal{M}}.
\end{equation}
Further implications of this substitution are discussed in the literature \cite{Ranftl2021b}. 
From here on we assume that the functional form of $\mathcal{M}$ is fixed, and we will suppress $\mathcal{M}$ in the conditional complex of the PDFs when not directly addressed to simplify notation, e.g.
\begin{equation}
    \p{\obs}{\xibf, \Xbf,  {\mathcal{D}}} \equiv \p{\obs}{\xibf, \Xbf,  {\mathcal{D}}, \mathcal{M}},
\end{equation}
and analogous to Eq.~(\ref{eq:def-sought-for-pdf}) we can then write
\begin{equation} \label{eq:aux2}
    \p{\obs}{\Xbf, \mathcal{D} \surryesno} = \int \p{\obs}{\xibf, \Xbf,{\mathcal{D}} \surryesno} \p{\xibf}{\hat{\Xbf}}\mathrm{d}V_{\xi}.
\end{equation}
Other popular functional forms of $\mathcal{M}$ are polynomial chaos expansions \cite{Xiu2005a,Crestaux2009a}, Gaussian process regressors \cite{OHagan1978a,Rasmussen2006a} or recently physics-informed NNs \cite{Tripathy2018a,Zhu2018a}. 
In Section~\ref{sec:nn-surrogates} NNs were introduced to construct a single surrogate model to learn and predict stresses at each node in $\hat{\Xbf}$ collectively, rather than a set of $N_{\rm x}$ surrogates that predict the stress at $\Xbf^{(n_{\rm x})},\;n_{\rm x}=1,\ldots,N_{\rm x}$ each.

\subsubsection{Approximation error of the surrogate}
In (\ref{eq:aux2}), we assumed that we can approximate the uncertainty of the Cauchy stress component $\obs$ by replacing the FE analysis with a surrogate learned from a training data set.  
This approximation introduces additional uncertainties via the surrogate parameters $\wbf$ through the first term under the integral in (\ref{eq:aux2}), i.e.
\begin{equation} \label{eq:marg-w-aux}
   \p{\obs}{\xibf, \Xbf, {\mathcal{D}} \surryesno} = \int \p{\obs}{\xibf, \xbf, \wbf, \cancel{{\mathcal{D}}} \surryesno} \p{\wbf}{{\mathcal{D}} \surryesno} \mathrm{d}V_{\mathrm{w}},
\end{equation}
where $\p{\wbf}{{\mathcal{D}} \surryesno}$ is the posterior PDF for the surrogate parameters. 
Note that the data ${\mathcal{D}}$ in the first term is superfluous, since $\obs$ is uniquely determined by given $\xibf$, $\Xbf$ and $\wbf$ via Eq.~(\ref{eq:def-surrogate}).
Substituting Eq.~(\ref{eq:marg-w-aux}) into Eq.~(\ref{eq:aux2}) yields
\begin{equation} \label{eq:marg-w}
    \p{\obs}{\Xbf, {\mathcal{D}}} = \iint \p{\obs}{\xibf, \Xbf, \wbf}\p{\wbf}{\mathcal{D}}\p{\xibf}{\Xbf}\mathrm{d}V_{\xi}\mathrm{d}V_{\mathrm{w}}.
\end{equation}
So far we have implicitly assumed that (\ref{eq:def-sought-for-pdf}) is approximated by (\ref{eq:aux2}), i.e. $\p{\obs}{\xbf, \mathcal{D}, \mathcal{M}}  \approx \p{\obs}{\Xbf} $. 
This is only valid if the surrogate model is actually a good approximation to the FE simulation, see (\ref{eq:def-surrogate}), and if at the same time the posterior for the surrogate parameters shows a sharp peak at the optimal surrogate parameters $\wbf^{\ast}$, as in (\ref{eq:optimization-problem}), $\p{\wbf}{{\mathcal{D}} \surryesno} \approx \delta(\wbf - \wbf^{\ast})$, see \cite{Ranftl2021b}. 
Consequently, we wrongly assumed that there are optimal surrogate parameters $\wbf^{\ast}$ that have no uncertainty.
In the following we keep the assumption (\ref{eq:stochastic-integration-brute-force}) and refrain from neglecting the uncertainties in the surrogate parameters $\wbf$ introduced by using a surrogate in the first place (\ref{eq:def-surrogate}). To put it another way, we aim to solve (\ref{eq:marg-w}). 
To do this, we must first define the remaining ingredients, i.e. likelihood and prior for the posterior of the surrogate parameters $\p{\wbf}{{\mathcal{D}} \surryesno}$. 

\subsubsection{Likelihood and prior} \label{sec:likelihood-prior}
Using Bayes' theorem and conditionally independent data from the FE analysis and recalling (\ref{eq:def-data}) we find that 
\vspace*{-0.2cm}
\begin{eqnarray}
    \p{\wbf}{{\mathcal{D}} \surryesno} & \propto & \p{\obsbigset}{\bm{\Xi}, \hat{\Xbf}, \wbf \surryesno} \p{\wbf}{\cancel{\bm{\Xi}}, \cancel{\hat{\Xbf}}  \surryesno} \label{eq:posterior-aux0}\\
    &=& \prior{\wbf}{} \prod_{n_{\xi}=1}^{N_{\xi}}\p{\obssmallset}{\xibf^{(n_{\xi})}, \hat{\Xbf}, \wbf \surryesno} \label{eq:posterior-aux} \\ 
    &=& \prior{\wbf} \prod_{n_{\xi}=1}^{N_{\xi}} \prod_{n_{\rm x}=1}^{N_{\rm x}} \p{\obs^{(n_{\rm x},n_{\xi})}}{\xibf^{(n_{\xi})}, \Xbf^{(n_{\rm x})}, \wbf \surryesno}. \label{eq:posterior-aux2}
\end{eqnarray}
In (\ref{eq:posterior-aux0}) it is recognized that surrogate parameters $\wbf$ are {\em a priori} conditionally independent of locations $\hat{\Xbf}$ and random fields $\xiset$, which will be discussed later. 
In (\ref{eq:posterior-aux}) it is assumed that the simulation output $\obssmallset$ corresponding to a specific random field sample $\xibf^{(n_{\xi})}$ provides no information about the simulation output corresponding to any other random field sample, i.e. conditional independence. 
Finally, in (\ref{eq:posterior-aux2}) we assumed that to measure the stress at a particular location for a given random field sample $\xibf^{(n_{\xi})}$ and surrogate parameters $\wbf$, we only have to specify this particular measurement location, namely $\Xbf^{(n_{\rm x})}$, and not all, $\hat{\Xbf} =\{\Xbf^{(n_{\rm x})}\}$.
This conditional independence in (\ref{eq:posterior-aux2}) might be counter-intuitive because one would think that if the random field $\xibf$ is spatially correlated through $\Xbf$ then the Cauchy stress components $\obs$ should also be spatially correlated. This is indeed true. However, as soon as $\xibf^{(n_{\xi})}$, $\Xbf^{(n_{\rm x})}$, and $\wbf$ are determined in (\ref{eq:posterior-aux2}) then $\obs^{(n_{\rm x},n_{\xi})}$ is uniquely determined by $\mathcal{M}$, as in (\ref{eq:def-surrogate}).
Then, the PDF for $\obs$ in (\ref{eq:posterior-aux2}) describes only the additional approximation error or uncertainty caused by the surrogate approximation (\ref{eq:def-surrogate}). 
This approximation error is assumed to be location-independent and therefore the same for all locations. 
Spatially correlated noise could be introduced at this point by another Gaussian process, for example. 
Also note that the surrogate model predicts stresses $\mathcal{M} = \{\mathcal{M}^{(n_{\rm x})}\}_{n_{\rm x}=1}^{N_{\rm x}}$ at all locations $\Xbf^{(n_{\rm x})}\in \hat{\Xbf}$ simultaneously, as mentioned in Section~\ref{sec:nn-surrogates}. 
The spatial correlations of the stresses $\obs$ are then implicit in the surrogate. 
As shown in Section~\ref{sec:nn-surrogates}, it is the particular advantage of NNs as a surrogate to enable learning and prediction of stresses at a large number of measurement sites together.

Due to the lack of information about the distribution for the simulation data, we assume a general exponential likelihood like a Gaussian as a convenient default choice for the likelihood presented in (\ref{eq:posterior-aux2}), more precisely for the likelihood $\p{\obs^{(n_{\rm x},n_{\xi})}}{\xibf^{(n_{\xi})}, \Xbf^{(n_{\rm x})},  \wbf \surryesno}$. 
Let ${\mathcal{L}}^{(n_{\rm x},n_{\xi})}$ be a loss function or mismatch term for a given data point $(n_{\rm x}, n_{\xi})$. 
Then, given the noise variance $\Delta^2$ that describes the uncertainty scale of the surrogate, the properly normalized likelihood for a single data reads
\begin{equation} \label{eq:single-likelihood}
        \p{\obs^{(n_{\rm x},n_{\xi})}}{\xibf^{(n_{\xi})}, \Xbf^{(n_{\rm x})}, \wbf, \Delta \surryesno} =  (2\pi \Delta^2)^{- 1/2} \exp \Big({-\frac{ {\mathcal{L}}^{(n_{\rm x},n_{\xi})}}{2\Delta^2}}\Big). 
\end{equation}
With the same argument as in (\ref{eq:posterior-aux0})-(\ref{eq:posterior-aux2}) and the total loss $\mathcal{L}$ the joint likelihood of all data given surrogate uncertainty $\Delta^2$ is
\begin{equation}\label{eq:joint-likelihood-given-Delta}
    \p{\obsbigset}{ \xiset, \hat{\Xbf}, \wbf, \Delta \surryesno} = (2\pi \Delta^2)^{- (N_{\rm x}N_{\xi})/2} \exp \Big({-\frac{{\mathcal{L}}}{2\Delta^2}}\Big),
\end{equation}
with
\begin{equation}
    \quad {\mathcal{L}} := \sum_{n_{\rm x}=1}^{N_{\rm x}} \sum_{n_{\xi}=1}^{N_{\xi}} {\mathcal{L}}^{(n_{\rm x},n_{\xi})},
\end{equation}
where $N_{\rm x}$ is the number of pivot points or the number of measurement sites in $\hat{\Xbf}$ while $N_{\xi}$ is the number of random field samples.
It follows that the number of single stress measurements is $N_{\rm x} N_{\xi}$.
Although the uncertainty scale of the surrogate  $\Delta^2$ is the same for all measurements, it is not known and must therefore be marginalized. 
We can reformulate the uncertainty as a precision parameter $\tau : = 1/\Delta^2$.
A suitable and conjugate prior for the precision parameter $\tau$ is the gamma distribution $\p{\tau}{a_1,b_1} = p_{\Gamma}(a_1,b_1)$, which leads to a Student-t distribution \cite{Murphy2007a} of the form
\begin{eqnarray}\label{eq:marginal-joint-likelihood}
    \p{\obsbigset}{ \xiset, \hat{\Xbf}, \wbf \surryesno}  &=& \int \p{\obsbigset}{ \xiset, \hat{\Xbf}, \wbf,  \tau \surryesno} \p{\tau}{a_0,b_0} \mathrm{d}\tau \\
    &=& (2\pi b_1)^{-(N_{\rm x}N_{\xi})/2} \frac{\Gamma\big[a_1+(N_{\rm x}N_{\xi})/2\big]}{\Gamma(a_1)}  \Big(1+\frac{{\cal{L}}}{2 b_1} \Big)^{-a_1-(N_{\rm x}N_{\xi})/2},
\end{eqnarray}
where $a_1 = 2$ and $b_1 = 2\cdot10^{-6}$ are chosen in analogy to \cite{Zhu2018a}. In order to fully define the likelihood, we now only have to specify the loss function ${\mathcal{L}}^{(n_{\rm x},n_{\xi})}$.
In this study, a smooth ${\mathcal{L}}_1$ loss, as in a Huber loss, does not lead to appreciably different results and is also more difficult to handle because it introduces another hyperparameter. Here we decided to use the ${\mathcal{L}}_2$-norm as the loss function, i.e. a Gaussian likelihood, which led to reasonable results. Formally this is
\begin{equation}
    {\mathcal{L}}^{(n_{\rm x},n_{\xi})} =  \Big[\obs^{(n_{\rm x},n_{\xi})} - \mathcal{M} (\xibf^{(n_{\xi})}; \wbf \mid \Xbf^{(n_{\rm x})} ) \Big]^2. \label{eq:loss-function} 
\end{equation}

For the prior we choose a multivariate Gaussian with a diagonal covariance matrix, i.e. $\p{\wbf}{\alpha \surryesno} = {\mathcal{N}}(0,\alpha^{-1} \mathbbm{1})$ which implies that the components of $\wbf$ are conditionally independent for a given {\em a priori} precision hyperparameter $\alpha$. With a gamma-type hyper-prior $\prior{\alpha} = p_{\Gamma}(a_0, b_0)$, we get a centered Student-t distribution \cite{Murphy2007a}, as 
\begin{eqnarray}
    \prior{\wbf} & = & \int \p{\wbf}{\alpha} \prior{\alpha} \mathrm{d}\alpha
     =  (2\pi b_0)^{-\frac{N_{\mathrm{w}}}{2}} \frac{\Gamma\big(a_0+N_{\mathrm{w}}/2\big)}{\Gamma(a_0)}  \Big(1+\frac{\wbf^2}{2 b_0} \Big)^{-a_0-N_{\mathrm{w}}/2},
\end{eqnarray}
where $N_{\mathrm{w}}$ is the number of surrogate parameters $\wbf$.
This choice regularizes against outliers and promotes sparsity in the weights $\wbf$. For the hyperparameters we choose $a_0 = 1$ and $b_0=0.05$, see \cite{Zhu2018a}.
We have now determined the un-normalized posterior in (\ref{eq:posterior-aux0}). 

The last missing ingredient, i.e. the first term under the integral in (\ref{eq:marg-w}), is already implicitly defined as the likelihood (see Eq.~(\ref{eq:single-likelihood})) for a single new data point, which in turn is integrated with respect to $\Delta^2$. Explicitly, this is
\begin{align} \label{eq:likelihood-single-new-data}
    \p{\obs}{\xibf, \Xbf, \wbf \surryesno} &=  (2\pi b_1)^{-1/2} \frac{\Gamma(a_1+1/2)}{\Gamma(a_1)}\bigg(1 + \frac{\big(\sigma-\mathcal{M} (\xibf; \wbf \mid \Xbf)\big)^2}{2b_1}\bigg)^{-a_1-1/2}.
\end{align}
We have now fully specified (\ref{eq:marg-w}) and will try to solve it below.

\subsubsection{Variational inference} \label{sec:variational-inference}
For generalized linear models $\mathcal{M}$, i.e. linear in the surrogate parameters $\wbf$, the integral (\ref{eq:marg-w-aux}) can often be solved analytically \cite{Ranftl2021b}. However, the linearity limits the expressive capacity of the surrogate model. 
In this study, we want to introduce an NN as a surrogate that has a nonlinear dependence on the surrogate parameters $\wbf$. The NN has $70\,020$ parameters, leaving us with a high-dimensional integral defined in (\ref{eq:marg-w}). This integral is difficult to solve numerically for exact inference, even with the most sophisticated variants of Markov Chain Monte Carlo currently available, and difficult to solve with Nested Sampling \cite{Skilling2006a}. So we approximate (\ref{eq:marg-w}) by variational inference \cite{Blei2017a}. In other words, the PDF $\p{\wbf}{{\mathcal{D}} \surryesno}$ from (\ref{eq:marg-w}) is approximated with a suitable PDF $\mathrm{q} \in Q$ from a family of functions $Q$, so that the integration with respect to $\wbf$ in (\ref{eq:marg-w}) becomes manageable. Then, $\mathrm{q}$ must be chosen such that the Kullback-Leibler divergence $\mathcal{K}(\mathrm{q};p)$ becomes minimal, i.e.
\begin{equation}\label{eq:var-inf-approx}
    \p{\wbf}{{\mathcal{D}} \surryesno}  \approx \mathrm{q}^{\ast}(\wbf ) = \arg \min_{\mathrm{q}\in Q} \mathcal{K}(\mathrm{q}; p),
\end{equation}    
with 
\begin{equation}\label{eq:KL-divergence}
    \mathcal{K}(\mathrm{q}; p) = \int \mathrm{q}(\wbf ) \log{\bigg[ \frac{\mathrm{q}(\wbf)}{\p{\wbf}{{\mathcal{D}} \surryesno}} \bigg]} \mathrm{d}V_{\mathrm{w}}. 
\end{equation}
The Kullback-Leibler divergence is a measure of the distance between two PDFs, namely $p$ and $\mathrm{q}$. 
This optimization problem can be solved by parameterizing $\mathrm{q}$ with tuning parameters $\theta$ such that $\mathrm{q} := \mathrm{q}_{\theta}$. 
The PDF $\mathrm{q}_{\theta} \in Q$ could then be the exponential family, e.g., Gaussian distributions with mean and variance $\theta :=(\mu, \varsigma^2)$.
Then analytical expressions for the gradient of the Kullback-Leibler divergence (\ref{eq:KL-divergence}) are often available, but with the disadvantage that one is limited to a family of parameterizable distributions.
A more advanced approach is Stein variational gradient decent \cite{Liu2016a,Liu2016b,Liu2017a,Zhu2018a}, which represents $\mathrm{q}$ numerically with samples and instead parameterizes small perturbations of $\mathrm{q}$ by small, parameterized coordinate transformations $T$ according to
\begin{equation}
    T(\wbf) = \wbf + \varepsilon \phibf(\wbf),
\end{equation}
with small $\varepsilon$ and a vector-valued function $\phibf$. This in turn defines a perturbed PDF $\mathrm{q}_T$ given by
\begin{equation}
    \mathrm{q}_{T} (\wbf) = \mathrm{q}\big(T^{-1}(\wbf)\big) \det\, [\nabla_{\mathrm{w}} T^{-1}(\wbf)]. 
\end{equation}
Then the minimization in (\ref{eq:var-inf-approx}) corresponds to the minimization with respect to $\phibf$.
The estimate for the expected gradient of the Kullback-Leibler divergence $\mathcal{K}(\mathrm{q}_T, p)$ was derived in \cite{Liu2016a,Liu2016b, Liu2017a}, from which the iterative procedure results
\begin{equation} \label{eq:w-update}
    \wbf^{(n_{\mathrm{m}})}_{n_\mathrm{t}+1} = \wbf^{(n_{\mathrm{m}})}_{n_\mathrm{t}} + \varepsilon_{n_\mathrm{t}} \phibf^{\ast}(\wbf^{(n_{\mathrm{m}})}_{n_\mathrm{t}}), \quad n_\mathrm{t}=1,\ldots,N_\mathrm{t},
\end{equation}
with
\vspace*{-0.2cm}
\begin{equation} \label{eq:stein-estimate}
    \phibf^{\ast}(\wbf) \approx \frac{1}{{\cal{Z}}} \frac{1}{N_\mathrm{m}} \sum_{n_{\mathrm{m}}^\prime=1}^{N_{\mathrm{m}}} \bigg[\kappa(\wbf^{(n_{\mathrm{m}}^\prime)}_{n_\mathrm{t}}, \wbf) \nabla_{\mathrm{w}^{(n_{\mathrm{m}}^\prime)}_{n_\mathrm{t}}} \log{\p{\wbf^{(n_{\mathrm{m}}^\prime)}_{n_{\mathrm{t}}}}{{\mathcal{D}} \surryesno}} + \nabla_{\mathrm{w}^{(n_{\mathrm{m}}^\prime)}_{n_{\mathrm{t}}}} \kappa(\wbf^{(n_{\mathrm{m}}^\prime)}_{n_{\mathrm{t}}}, \wbf) \bigg],
\end{equation}
where $n_\mathrm{t}$ is the number of previous iterations, $\cal Z$ is a normalization constant that can be neglected for optimization purposes, and $\kappa$ is an appropriate kernel function. 
For our purposes, the kernel function is set to $\kappa(\zeta,\zeta^\prime) = \exp[-(\zeta-\zeta^\prime)^2 \log \vert N_\mathrm{m}\vert /H^2]$, where $H$ is the median of the pairwise distances between the current samples $\{\wbf^{(n_{\mathrm{m}})}_{n_{\mathrm{t}}}\}_{n_{\mathrm{m}}=1}^{N_\mathrm{m}}$, as advocated in \cite{Liu2016a}. 
The kernel function $\kappa$ used here should not be confused with the convolution kernel in the NN.
During training of the NN, updates of the learning rate $\varepsilon$ were performed using a particular optimization algorithm called ADAM \cite{Kingma2015a} and a cosine annealing learning rate schedule as described in \ref{app:details-architecture}.

The equations (\ref{eq:w-update}) and (\ref{eq:stein-estimate}) define an iterative procedure of small coordinate transformations of $\wbf$ and subsequent small perturbations of $\mathrm{q}$. This iterative procedure finally leads to a $\mathrm{q}^{\ast}$ in a sample representation that is optimal in the sense of Eq.~(\ref{eq:KL-divergence}). 
We can then approximate the integral with respect to $\wbf$ in (\ref{eq:marg-w}) by the weighted sum over the samples or particles $\wbf^{(n_{\mathrm{m}})}$. These samples represent $\mathrm{q}^{\ast}(\wbf)$ by the following relation
\begin{equation} \label{eq:var-inference-particle-representation}
    \mathrm{q}^{\ast}(\wbf) =  \sum_{n_{\mathrm{m}}=1}^{N_\mathrm{m}} \delta(\wbf - \wbf^{(n_{\mathrm{m}})}) W_{\mathrm{w}^{(n_{\mathrm{m}})}},
\end{equation}
where $W_{\mathrm{w}^{(n_{\mathrm{m}})}}$ is the normalized weight of the particle ${\wbf^{(n_{\mathrm{m}})}}$. More precisely, it is defined in this study as $W_{\mathrm{w}^{(n_{\mathrm{m}})}} = 1/N_\mathrm{m}$.

The integral with respect to the random field $\xibf$ still remains unsolved. We can approximate it by a discrete sum over the samples of $\xibf$ drawn from the distribution $\p{\xibf}{\hat{\Xbf}}$, as discussed for Eq.~(\ref{eq:stochastic-integration-brute-force}). Substituting (\ref{eq:stochastic-integration-brute-force}) and (\ref{eq:var-inference-particle-representation}) through (\ref{eq:var-inf-approx}) together with (\ref{eq:posterior-aux0}) in Eq.~(\ref{eq:marg-w}) then gives the final result
\begin{equation} \label{eq:final-result}
    \p{\obs}{\Xbf^{(n_{\rm x})}, {\mathcal{D}} \surryesno} \approx \sum_{n_{\rm s} =1}^{N_{\mathrm{s}}} \sum_{n_{\mathrm{m}}=1}^{N_\mathrm{m}} \p{\obs}{\xibf^{(n_{\rm s})}, \Xbf^{(n_{\rm x})}, \wbf^{(n_{\mathrm{m}})}, {\mathcal{D}} \surryesno} \, W_{\xi^{(n_{\rm s})}}\, W_{\mathrm{w}^{(n_{\mathrm{m}})}},
\end{equation}
where $W_{\xi^{(n_{\mathrm{s}})}} = 1/N_{\mathrm{s}}$ is according to Sections~\ref{sec:sampling-gaussian-fields} and \ref{sec:sampling-uniform-field}. 
The individual terms in these sums are the likelihood for new data (Eq.~(\ref{eq:likelihood-single-new-data})) as argued in Section~\ref{sec:likelihood-prior}. With the above substitutions and the definitions provided in Section~\ref{sec:likelihood-prior} this explicitly is
\begin{align}
    \p{\obs}{\xibf^{(n_{\rm s})}, \Xbf^{(n_{\rm x})}, \wbf^{(n_{\mathrm{m}})}, {\mathcal{D}} \surryesno} &=  (2\pi b_1)^{-1/2} \frac{\Gamma(a_1+1/2)}{\Gamma(a_1)}\bigg(1 + \frac{\mathcal{L}^{(n_{\mathrm{x}}, n_\xi)}}{2b_1}\bigg)^{-a_1-1/2}.
\end{align}

The surrogate predicts stresses with significantly reduced computational effort compared to the original FE analysis, presented in Section~\ref{sec:boundary-value-problem}.  
Once the NN has been trained as a surrogate model from $N_{\xi}$ training examples of random field inputs and corresponding stresses, we can predict the stresses of a much larger number $N_{\mathrm{s}}$ of new and unseen random field examples that finally improve the Monte Carlo estimate in (\ref{eq:final-result}).
The accuracy of this Monte Carlo approximation depends on the number of independent samples $N_{\mathrm{s}}$ with $\mathcal{O}(1/\sqrt{N_{\mathrm{s}}})$.

The uncertainties (\ref{eq:def-std}) derived from the PDF in (\ref{eq:final-result}) do also include the uncertainties of the NN itself. 
Note that this result can also be interpreted as an average over $n_{\mathrm{m}}=1,\ldots,N_\mathrm{m}$ different NNs with surrogate parameters $\wbf^{(n_{\mathrm{m}})}$ and weight $W_{\mathrm{w}^{(n_{\mathrm{m}})}}$, each NN with its own predictions for the stress distribution $\obs$ for all given random field samples $\xibf^{(n_{\mathrm{s}})}$. Also note that this approach makes it possible to investigate isolated network uncertainties for fixed parameter fields $\xibf$.

\section{Results}\label{s:4}
In this study, the NN was trained to predict the results of the FE analysis of the boundary-value described in Section~\ref{sec:boundary-value-problem}. A total of $10\,000$ samples of FE solutions were available for this. Of these, $4\,200$ were used as training samples and $800$ as validation samples. The remaining $5\,000$ samples were used for testing, i.e. as further ground truth solution to evaluate the generalization capabilities of the trained NN. 


After creating a surrogate model, training and adapting hyperparameters, and defining the architecture of the NN, the uncertainties of the surrogate model need to be quantified. Using the Cauchy stress component $\sigma_{33}$ as the quantity of interest, a Bayesian encoder-decoder was trained to approximate the mapping of random fields as input to the field of the Cauchy stress component $\sigma_{33}$ as output. In the analysis of the Cauchy stress tensor, the component $\sigma_{33}$ naturally turned out to be dominant in this particular boundary-value problem, since it is aligned with the loading direction of the uniaxial extension test. The other components were two orders of magnitude smaller. Attempts have also been made to learn all tensor components by one network together, but gave only noisy predictions and no meaningful results. 
\begin{figure}[h!]
    \centering
    \includegraphics[width=0.9\textwidth]{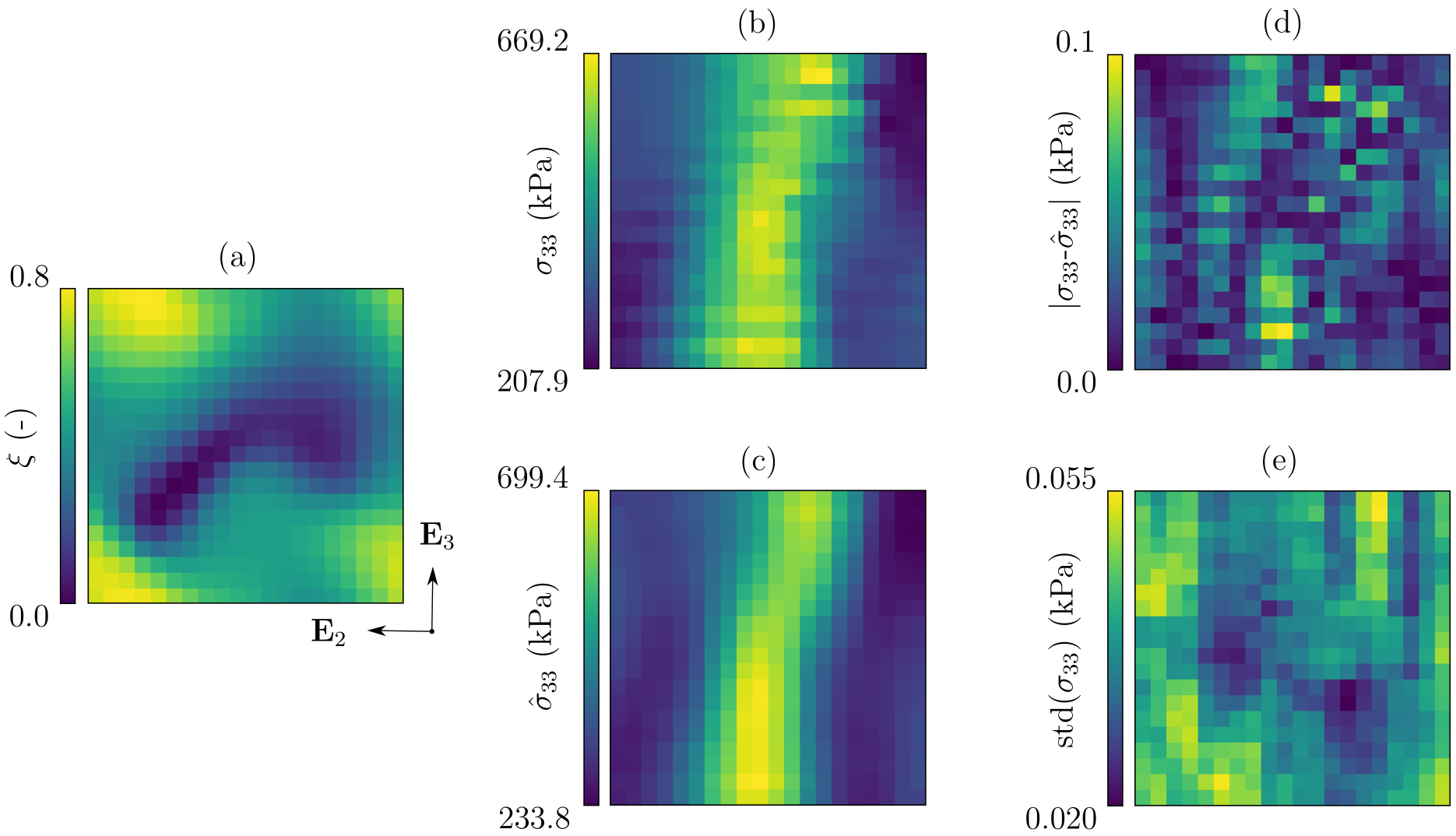}
     \vspace*{-0.1cm}
    \caption{Results of the trained NN or surrogate model: a representative random field (a) is chosen to compare the solution of the FE analysis or true target (b) with the mean prediction of the NN (c). The absolute difference between the mean prediction of the NN and the solution of the FE analysis is shown in (d), while the standard deviation of the NN prediction is shown in (e).}
    \label{fig-results_comparison}
\end{figure}

The input random field and the output of the FE analysis are shown in Figs.~\ref{fig-results_comparison}(a) and (b), respectively. The network prediction is plotted in Fig.~\ref{fig-results_comparison}(c) and the absolute difference between the output of the FE analysis and the NN prediction can be seen in Fig.~\ref{fig-results_comparison}(d). 
A comparison of the true output and its network prediction shows that the network cannot predict small-scale fluctuations within the data. Therefore, the network prediction looks pretty smooth compared to the Cauchy stress distribution obtained from the FE analysis. These fluctuations can be interpreted by the model as noise in the data and are ultimately responsible for the deviation of the NN prediction from the FE reference solution, as can be seen from the absolute difference in Fig.~\ref{fig-results_comparison}(d) and the predicted standard deviation in Fig.~\ref{fig-results_comparison}(e). Even if the network prediction seems to capture the output well at first glance, it is the elusive small-scale fluctuations that end up causing relative errors of up to $20$\,\%.
\begin{figure}[b!]
    \centering
    \includegraphics[width=0.9\textwidth]{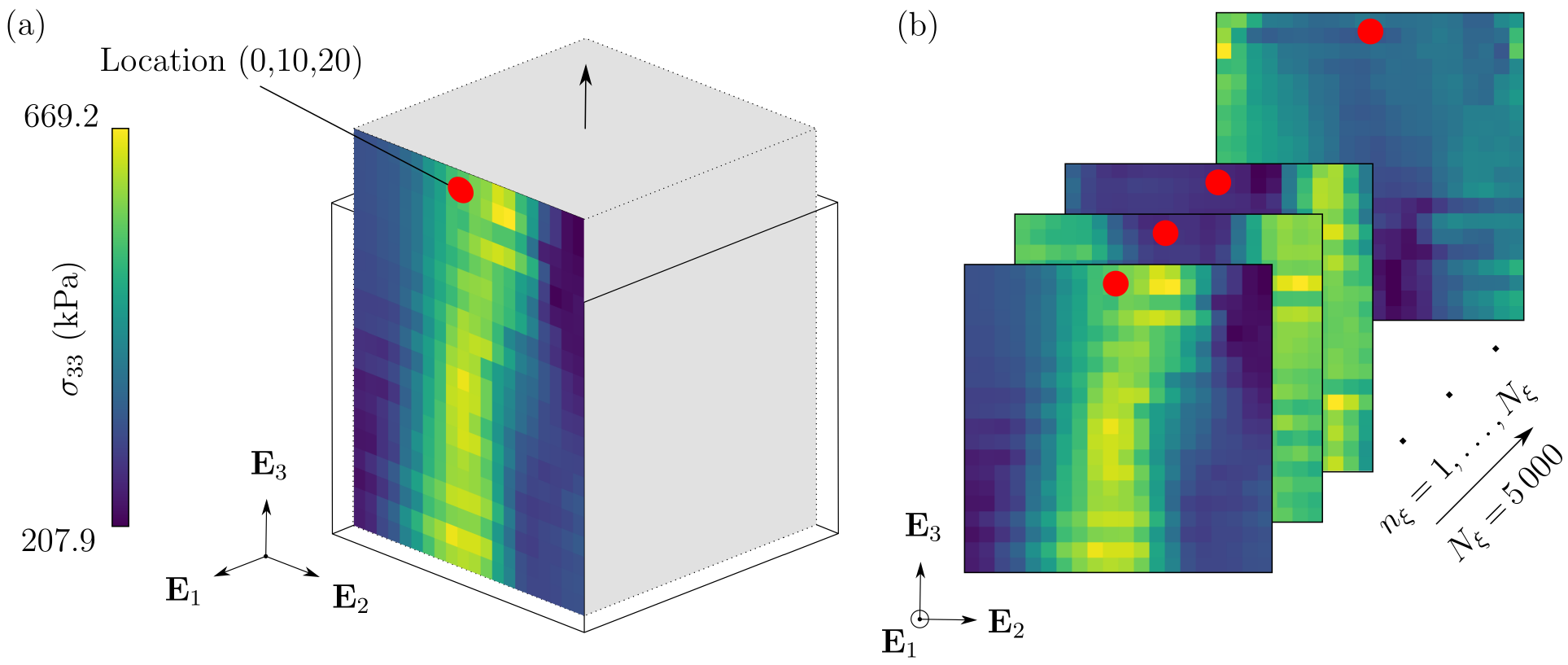}
    \vspace*{-0.25cm}
    \caption{
    Illustration of the calculation of the posterior probability density functions for the Cauchy stress (Eq.~(\ref{eq:final-result})) where the representative location (marked red) for evaluation is chosen at $(0, 10, 20)$: in (a) the output of the FE solution $\obssmallset$, i.e. the spatial distribution of the Cauchy stress component $\sigma_{33}$ for a given random field sample $\xibf^{(n_\xi)}$ is shown; (b) shows for each location the values $\sigma_{33}^{(n_\xi)}$ corresponding to each $\xibf^{(n_\xi)}$ of the $N_{\xi} = 5\,000$ samples, which are then aggregated into a histogram (Fig.~\ref{fig-results_UQ}). 
    This histogram obtained from the FE solution (see (a)) is used as a reference to compare with the NN. The posterior distribution predicted by the NN is formed analogously by aggregating, for each of the $N_{\mathrm{s}}$ random field samples $\xibf^{(n_{\mathrm{s}})}$, the corresponding values $\sigma_{33}^{(n_{\mathrm{s}})}$, as predicted by the NN. 
    For the NN prediction we also need to aggregate for each $\xibf^{(n_{\mathrm{s}})}$ the  $n_{\mathrm{m}} =1,\ldots,N_{\mathrm{m}}$ distinct predictions provided by the ensemble of $N_{\mathrm{m}}$ NNs (not shown here).}
    \label{fig-results_mean_prediction}
\end{figure}
\begin{figure}[h!]
    \centering
    \includegraphics[scale=1]{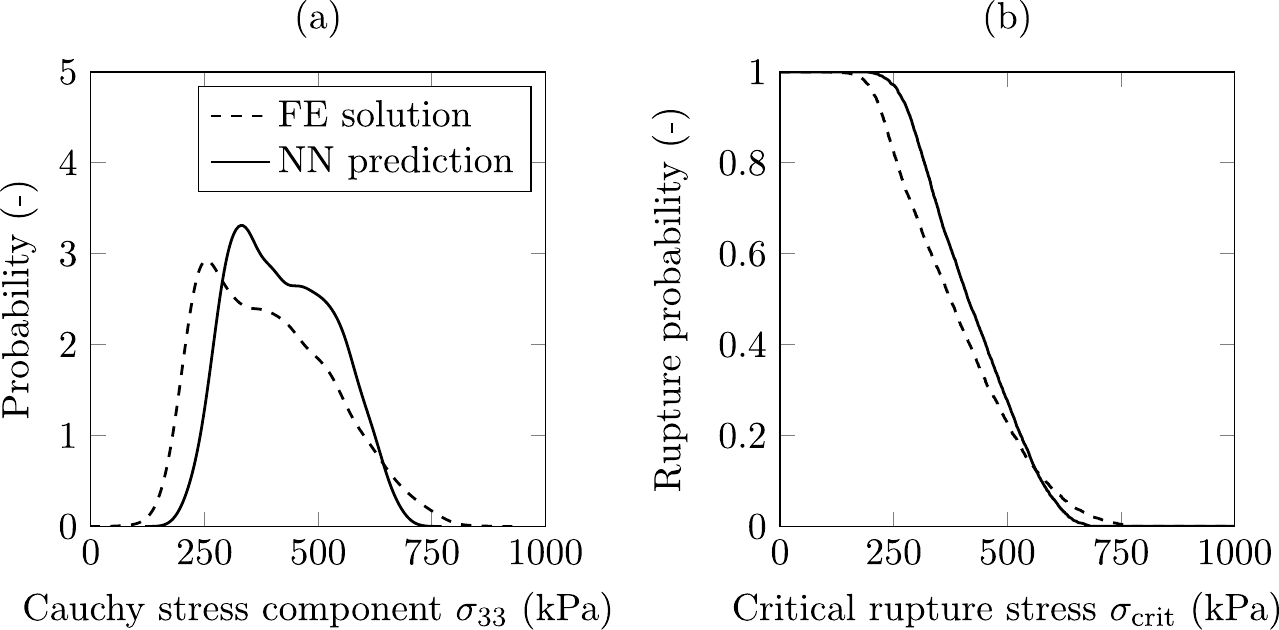}
    \caption{
    The two diagrams illustrate the posterior distribution for the Cauchy stress component $\sigma_{33}$ at the representative location $(10, 20)$ in the ($\Ebf_2,\Ebf_3$) plane: (a) posterior for all mean predictions; (b) inverse cumulative distribution of the posterior for all mean predictions compared to the FE solution, or the true target, respectively.}
    \label{fig-results_UQ}
\end{figure}

In order to quantify uncertainties, it is useful to compare surrogate predictions at certain locations with the true solution, here the FE analysis, as illustrated in Fig.~\ref{fig-results_mean_prediction}. For this we have chosen the representative location $(0, 10, 20)$, which lies in the ($\Ebf_2, \Ebf_3$) plane. 
Figure~\ref{fig-results_UQ}(a) then shows the posterior probability distribution for the Cauchy stress component $\sigma_{33}$ at this representative location, calculated from the FE reference and the NN. In total $5\,000$ NN predictions were compared to the respective output of the FE analysis. 
After training with the first $N_{\xi}=$ $5\,000$ samples, $N_{\mathrm{s}} = 5\,000$ predictions of the invisible Cauchy stress component $\sigma_{33}$ were made and their frequency plotted in a histogram. Comparing the solution of the FE analysis with the NN prediction, it is noticeable that the properties of the two curves are similar, including the shoulder on the right hand side. However, the NN is too confident, i.e. the reference distribution of the FE solution is wider. This is again due to the small-scale fluctuations. Furthermore, in this study we compared the uncertainties of the FE analysis and the NN at several positions with similar results.

Figure~\ref{fig-results_UQ}(b) displays the inverse cumulative distribution of the Cauchy stress component $\sigma_{33}$, which can be interpreted as a rupture probability for the aortic wall. In other words, at a given critical Cauchy stress value, above which the aortic tissue is likely to fail, the surrogate model can be used to support the FE results fairly  accurately. Note that the high gradient in some regions can introduce significant biases in the rupture probability. Similar to Fig.~\ref{fig-results_UQ}(a), the properties of the two curves are similar and an offset can be seen. The test results for the NN indicated an accuracy of $86$\,\%, as shown in the Appendix (Fig.~\ref{fig:reliability-diagram}).  

\section{Discussion}\label{s:5}
We have presented a stochastic approach to model material inhomogeneities in the aortic wall using the example of aortic dissection. To describe pathological changes in the aortic wall, a constitutive framework was introduced that includes a degradation parameter to model degraded elastic fibers. This parameter was then assumed to be spatially distributed within the aortic wall. Based on this assumption, a beta random field of the degradation parameter was developed and sampled. Subsequently, the stochastic constitutive model was implemented in FEAP~\cite{FEAP} and applied to a boundary-value problem, more precisely a uniaxial extension test, which provided the stress distribution as the quantity of interest in this study. 

However, the results of the stochastic model are meaningless without accounting for the uncertainties introduced. In this study, uncertainty propagation was achieved using a NN as a surrogate model. This approximation introduces an additional uncertainty that can be treated in the context if parametric input uncertainties on the basis of Bayesian probability theory. The additional uncertainty, i.e. the uncertainty of the network itself, was estimated using a variational inference formulation. The variational inference limitation of parametrized PDFs is overcome with a particle representation of the approximated PDF. This effectively results to an ensemble of NNs whose predictions are averaged. In addition to the degradation parameter uncertainties, other model parameter uncertainties were not considered, e.g., uncertainties in the strain-energy function (\ref{eq:isochoric_psi}). Uncertainties in geometry and boundary conditions could be neglected in this virtual laboratory setting, but can be important in more physiological models, e.g., patient-specific models. Moreover, the numerical accuracy of the FE solver, e.g., the discretization of domains, random fields, fiber distributions or NN can be neglected in view of the model uncertainty.

At this point, the discrete character of the applied DFD method must be discussed with regard to the uncertainties of the model. As emphasized in previous studies \cite{RolfPissarczyk2021a, RolfPissarczyk2021b}, the exclusion of degraded elastic fibers strongly depends on the number of discrete elements on the unit hemisphere. 

For reasons of computational time, we decided to carry out the FE analysis with $m=640$ spherical triangles. A finer mesh would have made the constitutive model more sensitive to small changes in the degradation parameter. In other words, if the change in the degradation parameter is smaller than the discrete steps between two spherical triangles, then the model may not accurately reproduce these changes in the degradation parameter. A non-uniform discretization or a higher number of elements could reduce the uncertainties of the constitutive model. However, a significant increase in the number of elements would result in an unfeasible increase in computational time.

A comparison of the NN-based approximation of the PDF for the Cauchy stress component $\sigma_{33}$ with the reference solution (Fig.~\ref{fig-results_UQ}) showed that the NN delivers a qualitatively similar result, but with the uncertainties underestimated. 
This effect is probably due to the problem that the NN could not predict small-scale fluctuations in the stresses, as mentioned in Section~\ref{s:4}. These small-scale fluctuations can more frequently result in stress values that are lower or higher than the smoothened \lq mean field' prediction of the NN.
More precisely, the NN interprets the small-scale fluctuations as noise or cannot distinguish between noise and small-scale fluctuations. Since we are trying to avoid over-fitting or fitting of the noise while training the NN, it is unlikely that this problem will be solved simply by using more data.
In our Bayesian approach, each prediction of the NN is actually the average of the predictions of an ensemble of $N_\mathrm{m}=20$ NNs, see Eq.~(\ref{eq:final-result}), each with their own set of optimal weights and biases. This averaging effectively results in smoothened predictions. However, when comparing the individual predictions in the ensemble of NNs for a given random field input, we found that the individual NNs in the ensemble already suffered from the inability to predict small-scale fluctuations. A regularization of the training with a Huber loss function, as discussed for Eq.~(\ref{eq:loss-function}), did not bring notably better results.

The small-scale fluctuations can possibly be captured with modifications to the model, which we will discuss below. One possibility would be to dispense with the conditional independence of the noise at different locations, i.e. the product form in the likelihood, see Eq.~(\ref{eq:posterior-aux2}). Instead, the small-scale fluctuations could be modeled explicitly by imposing a Gaussian process prior on $\p{\obssmallset}{\xibf^{(n_{\xi})}, \hat{\Xbf}, \wbf \surryesno}$, i.e. correlating the NN uncertainty at different locations, possibly with a non-smooth Mat{\'e}rn-class covariance function.

Another possibility would be to adapt the architecture so that the small-scale fluctuations are easier to learn. However, the complex architecture of the network complicates the  interpretation of the network, and it is not obvious what this modification must look like. For example, simpler architectures with a physics-informed NN \cite{Raissi2018a,Raissi2019a} are promising and require less data for training, but did not provide useful results in the present study. Possible reasons are the fact that the physics is governed by the entire stress tensor and not just by a specific component of the stress tensor, which indicates an  inconsistent training objective \cite{Rohrhofer2021a}. 
Attempts to learn all the stress tensor components jointly suffered again from the  indistinguishability of small-scale fluctuations and noise with both the introduced encoder-decoder and the physics-informed NN.
Another reason could be that boundary conditions were not taken into account in the network. 
This issue can be resolved by a soft or hard enforcement of boundary conditions \cite{Gao2020a}. 
Another direction would be to equip the architecture with attention mechanisms, as in natural language processing transformers \cite{Vaswani2017a} or vision transformers \cite{Dosovitskiy2021a}. While our dense blocks pass forward all features of all scales to subsequent layers, a transformer could allow to automatically weight and correlate features on multiple scales. This could reduce the number of network parameters and allow for more data-efficient training while retaining the important feature correlations across multiple scales, at the expense of a loop in the backpropagation. This could make it easier to find a set of weights and biases that accurately predict small-scale fluctuations. Unfortunately, the convergence properties of the discussed NNs are poorly understood \cite{LeCun2015a,Goodfellow2016a}.

Note that Gaussian processes {\em inter alia} suffer from the curse of dimensionality in the input space, while NNs shift the curse to the weight space. In this context, this is the particular advantage of NNs, since the dependency of the simulation on a parameter field can be learned with it. If the number of degrees of freedom in the FE model scales linearly with the number of nodes $N_{\rm x}$, then the inversion of the stiffness matrix scales as $\mathcal{O}(N_{\rm x}^3)$. Although several much faster methods exist, the following argument still holds. The total computation time for the full problem scales as $\mathcal{O}(N_{\rm s}N_{\rm x}^3)$ for a brute force approach.
When using a surrogate model, the total computation is composed of (i) the computation time for generating a training data set of $N_{\xi}$ examples, i.e. $\mathcal{O}(N_{\xi}N_{\rm x}^3)$, (ii) the computation time for the training and (iii) the computation time for $N_{\rm s}$ subsequent predictions with the trained network. The computational costs for the evaluation of new inputs in the trained NN (iii) depend on the respective architecture, but is negligible in most cases. This is the great advantage of any surrogate model. The bottleneck lies in the training time and data generation, and it is by no means clear how much training data $N_{\xi}$ and how much training time are necessary to produce a trustworthy NN in this context. In fact, for most architectures, there is no guarantee of training convergence at all, and it also depends on the architecture. In this study, data generation and training took about $33$\,h and $6$\,h, respectively. In contrast, the reference solution required about $66$\,h.

Adaption of the network to realistic geometries with irregular non-uniform meshes is conceptually easy with geodesic convolutions \cite{Masci2015a}, i.e. representing the convolution kernel in local coordinates, or graph NNs \cite{Zhou2020a}. The scaling of random field generation to large domains was addressed in \cite{Panunzio2018a,DeCarvalhoPaludo2019a}. 
However, practical limitations could again be the computational budget for data generation, i.e. $N_{\rm s}$ FE analyses and training. Liang et al.~\cite{Liang2018a} showed that it is possible to learn an accurate surrogate for the stress distribution on larger domains as a function of geometry using statistical shape models with far less data ($<800$), but neglecting surrogate uncertainties. It remains unclear whether the proposed approach is also really useful for random parameter fields on patient-specific geometries. In order to scale to patient-specific models, it would be promising to define the parameter field via a statistical shape model instead of directly via the FE discretization.

In principle, other machine learning approaches could have been used instead of a NN, e.g., a warped Gaussian process regressor \cite{Snelson2003a}. While the convergence properties and uncertainties of Gaussian processes are much better understood, they are unable to capture higher-order correlations, scaling to high-dimensional input data suffers from the curse of dimensionality, and big data sets also require approximations \cite{Hensman2013a}. Other machine learning approaches such as random forests \cite{Breiman2001a}, all too often suffer from the same limitations.

For the beta random field of the degradation parameter, we chose a correlation length of about $\sqrt{2}/3$\,mm, which is not based on experiments and is considered a limitation of this study. Therefore, future work has to investigate the sensitivity of the stress distribution and possible stress accumulations to the correlation length. In particular, recently published experimental results investigating the regional behavior of arterial samples tested {\rm in vitro} under physiologically relevant loading can be used \cite{Bersi2016a,DiGiuseppe2021a,Genovese2021a}. 

Regardless of the knowledge that many vascular diseases indicate local alterations in the aortic wall composition \cite{Bersi2016a}, most patient-specific computational models assume homogeneous material properties and a constant wall thickness \cite{Qiao2019a,Baeumler2020a}. In this study, we therefore presented a stochastic constitutive framework that provides a promising framework to study the role of material inhomogeneities using the example of spatially distributed degradation of elastic fibers under simplifying boundary conditions. This framework can also be applied and further extended to any other constitutive model. 
In order to be able to derive reliable conclusions about the stress distribution, future work must on the one hand contain the application to boundary-value problems that are closer to {\em in vivo} conditions and on the other hand the correlation length of the local inhomogeneities must correspond to the experiments \cite{Bersi2016a,DiGiuseppe2021a,Genovese2021a}. 
It is proposed that the presented Bayesian framework allows the identification of a law for stochastic inhomogeneities based on experimental data. We therefore recommend experimentally investigating the role of local alterations in the aortic wall and incorporating these regional changes in material properties in future models. In particular, modeling the correlation between regional pathological changes of different constituents could be of crucial importance.

\section*{Data and code availability}
Source codes for {\tt Python} (Section~\ref{s:2}) and for {\tt Python, PyTorch} (Section~\ref{s:3}) are available at URL: xxx (the file will be uploaded to {\tt https://repository.tugraz.at/}).

\section*{Funding}
This work was funded by Graz University of Technology, Austria through the Lead Project on the \lq Mechanics, Modeling, and Simulation of Aortic Dissection' (\url{biomechaorta.tugraz.at}) and supported by GCCE: Graz Center of Computational Engineering.

\section*{Acknowledgements}
The authors would like to acknowledge the use of HPC resources provided by the ZID of Graz University of Technology, Austria.

\section*{Conflict of interest}
We declare that we have no competing interests.

\appendix
 
\section{Proof of Eq.~(\ref{eq:beta-marginal})}\label{app:proof-beta-margin}
Let $\mathrm{f}_1, \mathrm{f}_2$ be two normal distributed, univariate random variables.
If $p(\mathrm{f}_1) = {\mathcal{N}}(\mu_1, \varsigma_1^2)$ and $p(\mathrm{f}_2) = {\mathcal{N}}(\mu_2, \varsigma_2^2)$, and $\mathrm{f}_1$ and $\mathrm{f}_2$ are independent, then we can write
\begin{equation}
    p(\mathrm{f}_1, \mathrm{f}_2) = (4\pi^2 \varsigma_1^2\varsigma_2^2)^{-1/2} \exp \bigg[-\frac{(\mathrm{f}_1-\mu_1)^2}{2\varsigma_1^2}\bigg] \exp\bigg[-\frac{(\mathrm{f}_2-\mu_2)^2}{2\varsigma_2^2}\bigg].
\end{equation}
Subsequently we know that $\mathrm{g}:=\mathrm{f}_1^2+\mathrm{f}_2^2$ follows a $\chi^2$-distributions. If $\mathrm{f}_1$ and $\mathrm{f}_2$ have the same variance, i.e. $\varsigma_1^2 = \varsigma_2^2 = \varsigma^2$, and the mean is defined as $\mu_1=\mu_2=0$, we get the special case of the gamma distribution $\Gamma(\eta,\gamma)$ with the parameters $\eta = 1/2$ and $\gamma = 1/2\varsigma^2$. For a sum of $2s$ squared Gaussian variables that all have the same variance, i.e. $\mathrm{g} = \sum_{r=1}^{2s} \mathrm{f}_r^2$, we find a gamma distribution with $\eta = s$ and $\gamma =1/2\varsigma^2$. This is a standard result and can, e.g., simply be shown by using characteristic functions.
Now let $\mathrm{g}_1$ and $\mathrm{g}_2$ be two independent such gamma-distributed variables, $p(\mathrm{g}_1) = \Gamma(\eta_1, \gamma_1)$ and $p(\mathrm{g}_2) = \Gamma(\eta_2, \gamma_2)$, with the parameters $\eta_1, \gamma_1$ and $\eta_2, \gamma_2$, respectively. Then the joint PDF reads
\begin{equation}
    p(\mathrm{g}_1, \mathrm{g}_2) = \frac{\gamma_1^{\eta_1}}{\Gamma(\eta_1)} \mathrm{g}_1^{\eta_1-1}\exp\,(-\gamma_1 \mathrm{g}_1)  \frac{\gamma_2^{\eta_2}}{\Gamma(\eta_2)} \mathrm{g}_2^{\eta_2-1}\exp\,(-\gamma_2 \mathrm{g}_2).
\end{equation}
Then with the variable transformation $\beta := \mathrm{g}_1/(\mathrm{g}_1 + \mathrm{g}_2)$ and $\lambda := \mathrm{g}_1+\mathrm{g}_2$ we obtain  $\mathrm{g}_1 = \lambda\beta$ and $\mathrm{g}_2 = \lambda(1-\beta)$. Thus, we find that
\begin{equation}
    p(\beta, \lambda) \mathrm{d} \beta \mathrm{d} \lambda
    = \frac{\gamma_1^{\eta_1} \gamma_2^{\eta_2}}{\Gamma(\eta_1) \Gamma(\eta_2)} \lambda^{\eta_1+\eta_2-1} \exp\,(-\gamma_1 \lambda\beta) \beta^{\eta_1-1}(1-\beta)^{\eta_2-1} \exp\,[-\gamma_2 \lambda(1-\beta)] \mathrm{d} \beta \mathrm{d} \lambda.
\end{equation}
We may now choose $\gamma_1 =\gamma_2=\gamma$, which results in $p(\beta,\lambda) = p(\beta)p(\lambda)$. Consequently, this gives  
\begin{equation}
    p(\beta) = \int p(\beta,\lambda) \mathrm{d}\lambda 
    \propto \beta^{\eta_1-1}(1-\beta)^{\eta_2-1},
\end{equation}
so $\beta := \mathrm{g}_1/(\mathrm{g}_1+\mathrm{g}_2)$ indeed follows a beta distribution with $\eta_1 = s$ and $\eta_2 = s^\prime$.
Next we have to extend these considerations from random variables $\beta$ to random fields $\betabf(\hat{\Xbf})$. With two Gaussian random fields $\fbf_1(\hat{\Xbf})$ and $\fbf_2(\hat{\Xbf})$ which have the same covariance function, and by defining with $\mathrm{f}_r^{(n_{\rm x})} = \mathrm{f}_r(\Xbf^{(n_{\rm x})})$ the value of random field $r=1,2$ at a particular location $\Xbf^{(n_{\rm x})}$, and by using the definition of a Gaussian process, we find that 
\begin{gather}
    \p{\fbf_r}{\hat{\Xbf}} = {\mathcal{N}}(0, \mathbf{K}(\hat{\Xbf}, \hat{\Xbf})) \nonumber
    \\ 
    \implies \p{\mathrm{f}_r^{(n_{\rm x})}}{\Xbf^{(n_{\rm x})}} = \int \p{\fbf_r}{\Xbf} \prod^{N_{\rm x}}_{\substack{n_{\rm x}^{\prime}=1\\
    n_{\rm x}^{\prime}\neq n_{\rm x}}}{\mathrm{d}\mathrm{f}_r^{(n_{\rm x}^{\prime})}} =  {\mathcal{N}}(0,\mathrm{k}(\Xbf^{(n_{\rm x})}, \Xbf^{(n_{\rm x})})),
\end{gather}
where the integral is with respect to all variables $\mathrm{f}_r^{(n_{\rm x}^{\prime})}$, $n_{\rm x}^{\prime} = 1,\ldots,N_{\rm x}$ except $\mathrm{f}_r^{(n_{\rm x})}$, while $\mathrm{k}$ and $\mathbf{K}$ are defined in (\ref{eq:def-kernel}) and (\ref{eq:def-discrete-gauss-random-field})$_2$. Since this holds $\forall r$ and $\forall \Xbf$ we reduced the problem to what was shown previously. Note that by definition of the Gaussian process, $\fbf_r(\Xbf)$ and $\fbf_r(\Xbf^\prime)$ are not independent, however, by definition of the procedure, $\fbf_r(\Xbf)$ and $\fbf_{r'\neq r}(\Xbf)$ are independent, which is sufficient.

In summary, we have first shown that a random variable $\beta$, as defined above, follows a beta distribution. Second, we have shown that this procedure also works for beta random fields by reducing the Gaussian random fields to Gaussian random variables at arbitrary locations. With this we have proven Eq.~(\ref{eq:beta-marginal}) and thus shown that the random field $\betabf$ constructed in Section~\ref{sec:sampling-uniform-field} is in fact a beta random field.

\section{Details of the encoder-decoder architecture}
\label{app:details-architecture}

\begin{figure}
	\centering
	\includegraphics[width=0.95\textwidth]{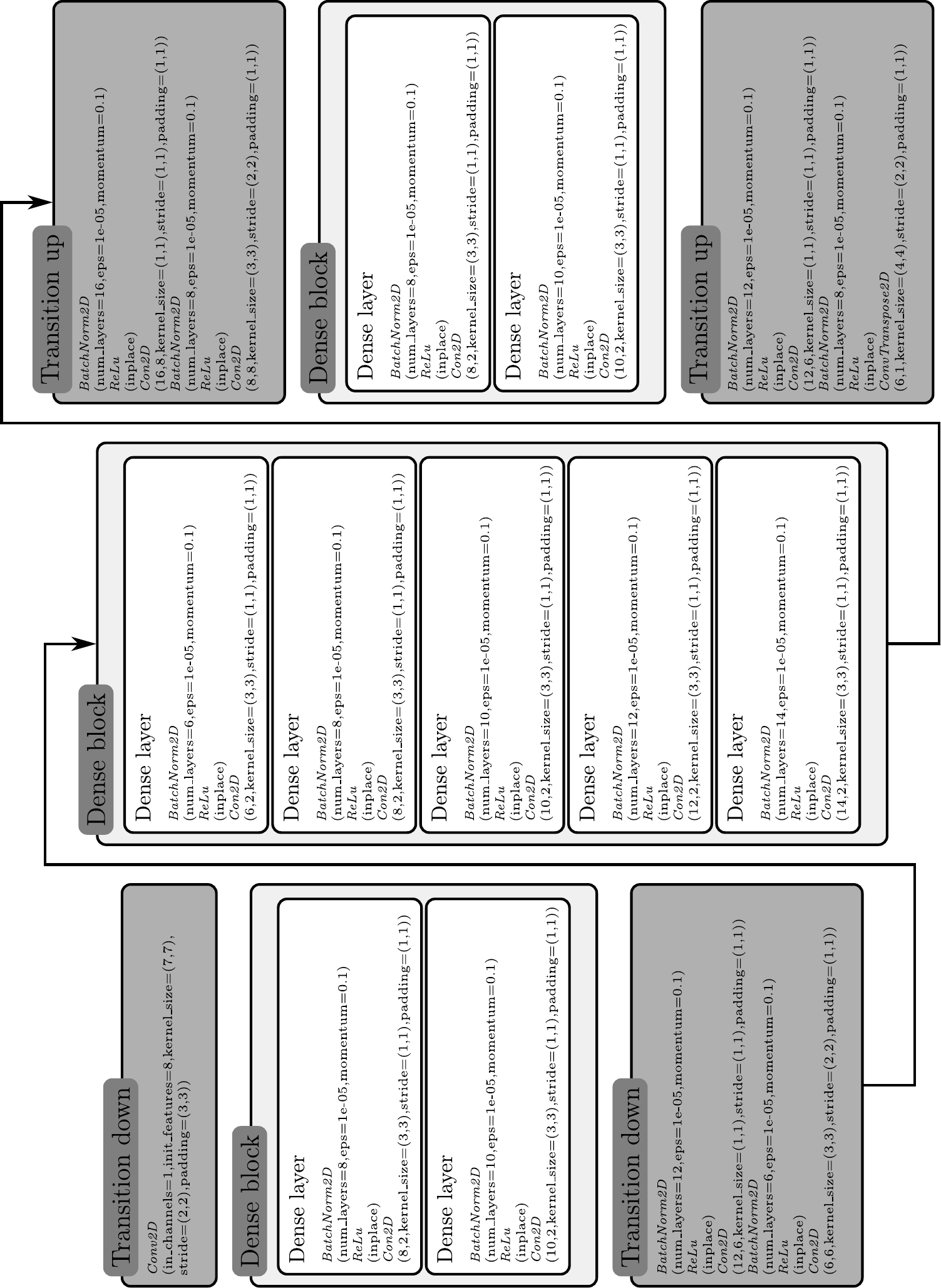}
	\caption{Technical illustration of the NN architecture. To ease re-implementation, we follow the nomenclature of PyTorch~\cite{pytorch_doc}. The first two parameters of the operation Conv2D (2D convolution) correspond to the number of input and output filters (channels).}
	\label{fig:architecture-technical-illustration}
\end{figure}

More information about the NN architecture can be found here.
The model was built and trained in Python 3.7.9 using Pytorch version 1.8.0 with CUDA 11.1. The generation of the data, i.e. FE simulation, was carried out on commercial desktop computers. The network has a total of $70\,020$ weights and biases. The model training and evaluation were performed on a server workstation with 12\,CPU cores (Intel Xeon E5-2630 v2), 128\,GB RAM, and 6 NVIDIA graphics processing units (GPUs) with 6\,GB GDDR5 memory each (NVIDIA Tesla K20Xm). The architecture is documented in detail in the technical illustration of Fig.~\ref{fig:architecture-technical-illustration}.
The associated architecture parameters and training parameters are also summarized in Tables~\ref{tab:NN_1} and \ref{tab:NN_2}. The main building blocks of the architecture are briefly described below. 

\begin{table}
	\parbox{.45\linewidth}{
		\centering
		\begin{tabular}{ll}
			\toprule
			Network property &  Value \\
			\midrule
			Batch size & $350$ \\     
			Dense blocks & $[2, 5, 2]$  \\
			Epochs  & $500$   \\
			Growth rate & $2$     \\  
			Learning rate & $0.03$ with cosine\\[-1.0ex]
			              & annealing    \\
			Number of single predictions\\[-1.0ex] for mean field predictions & $20$  \\
			Dense blocks encoder & $2$  \\
			Dense blocks decoder & $2$ \\
			Dense layers & $5$ \\
			Bottleneck size  & $1\times\mbox{growth rate}$  \\
			\bottomrule
		\end{tabular}
	\vspace*{-0.4cm}
		\caption{Properties, parameters of trained NN.}\label{tab:NN_1}
	}
	\hfill
	\parbox{.45\linewidth}{
		\centering
		\begin{tabular}{ll}
			\toprule
			Data property &  Value \\
			\midrule
			Total number & $10\,000$ \\
			Training set & $4\,200$ \\
			Test set & $800$ \\
			Validation set  & $5\,000$ \\
			Input size [px] & $20\times 20$ \\
			Output size [px] & $20 \times 20$\\
			\bottomrule
		\end{tabular}
		\vspace*{-0.15cm}
		\caption{Training data for the NN.}\label{tab:NN_2}
	}
\end{table}

\subsection{Convolution layer}
The convolutional layer is the fundamental element of a CNN. The activation function $\mathbf{h}$ of a  convolutional layer is, in its simplest form, a convolution characterized by a convolution kernel that is slided over the image pixel by pixel. The convolution kernel is usually expressed as a square matrix of size $h$, where $h$ defines the neighborhood of pixels or features with which each pixel is convoluted. For example, if $h=1$ then each pixel $P_{u,v}$ is only convoluted with itself. In contrary, with $h=3$ each pixel is convoluted with all nearest neighbor pixels. For higher $h$, each pixel is convoluted with the nearest neighbors, next nearest neighbors, next next nearest neighbors, and so on. Thus, the convolution kernel size defines the region in which a particular feature can be found. We can therefore think of it as a window matrix centered at $P_{u,v}$ and of size $h \times h$.
The stride $d$ is the number of pixels that the convolution window moves across the image on each iteration. If $d=1$, then a convolution with kernel matrix $h$ is computed for each pixel. If $d=2$ then every other pixel is skipped, reducing the number of feature maps and allowing to reduce the number of weights required and the corresponding memory and GPU requirements.
Padding refers to the addition of zero-valued pixels at the border of the image to ensure a well-defined convolution of pixels around the border.
In practice, the activation function of a convolutional layer is often a rectified linear unit (ReLU) function, defined as
\begin{equation}
    \mathrm{y}(\mathrm{x}) = \mathrm{max}\,(0, \mathrm{x}).
\end{equation}
The pixels corresponding to the respective convolution kernels are then found by optimizing the neural weights $\wbf$.

\subsection{Dense block}
A dense block is a basic module that directly connects all layers with one another. This implies that all interconnected layers have the same input and output dimensions. This idea was introduced by Huang et al.~\cite{Huang2017a} as \lq DenseNet'.
In other words, each layer $l$ is connected to all previous layers $\ell-1, \ell-2,\ldots$ in the same dense block. So if an image has $C_0$ input channels, e.g., for a RGB image $C_0 = 3$ each $\ell^{th}$ layer has a number of $C_0 + (\ell-1)C_{\rm gr}$ input feature maps $[\mathcal{M}^{(\ell-1)}]_{uv}$. Since we only learn the Cauchy stress component $\sigma_{33}$, it follows that $C_0 = 1$. 
However, the input features could correspond to all components of the Cauchy stress tensor, i.e. $C_0 = 9$. Two design parameters are introduced here, namely the number of layers within a dense block $M$ and the growth rate $C_{\rm gr}$. This defines the growth of the input feature maps for each layer as the number of features increases due to the connection to all previous layers. 
Then the total number of feature maps grows linearly with each layer introduced, so that a total of $C_{\rm out} = C_0 + M C_{\rm gr}$ feature maps are output. This also means that for a dense layer we need to modify (\ref{eq:network-recursive}) as follows
\begin{equation} 
    \mathcal{M}^{(\ell)} = \mathbf{h}_{\ell, n_\ell} \bigg( \bigoplus_{\ell^\prime<\ell}^{L}  \sum_{n_{\ell^\prime}=1}^{N_{\ell^\prime}} \wbf_{(\ell-1,n_\ell), (\ell^\prime-1, n_{\ell^\prime})} \odot \mathcal{M}^{(\ell^\prime-1)} + \bbf^{(\ell^\prime-1)} \bigg),
\end{equation}
%
where $\oplus$ is the concatenation of outputs from previous layers. The activation $\mathbf{h}$ is applied element-wise to the concatenated elements. The additional double index denotes the weight between neuron $n_\ell$ in layer $\ell$ and neuron $n_{\ell^\prime}$ in layer $\ell^\prime<\ell$. 

Image-to-image regression with encoder-decoder networks requires down-sampling and up-sampling to resize the feature maps, which makes concatenation of feature maps impossible. Therefore,  dense blocks and transition layers are introduced to solve this issue.

Similar to conventional CNNs, DenseNet includes batch normalization \cite{Ioffe2015a}, ReLU \cite{Glorot2011a}, convolution (Conv), and transposed convolution (ConvT), with padding for down-sampling and up-sampling, respectively, to ensure correct dimensions from the input feature map to a desired output feature map and {\em vice versa}.

\subsection{Transition layers}
Transition layers are used to reduce the number of feature maps between dense blocks and their size. More specifically, the encoding layer typically halves the size of feature maps, while the decoding layer doubles the size of the feature map. Both layers reduce the number of feature maps \cite{Zhu2018a}. 
In addition, batch normalization layers are used after each convolutional layer, since this can also be seen as an effective regularizer \cite{De2020a}.

As proposed in \cite{Long2015a}, fully convolutional networks are the extension of CNNs for pixel-to-pixel predictions, where fully convolutional networks replace the fully connected layers of CNNs with convolutional layers. Furthermore, up-sampling layers are added at the end to restore the input spatial resolution, and skip connections between feature maps are included for the down-sampling and up-sampling path, see \cite{Zhu2018a}.

This work adopts the architecture of \cite{Zhu2018a}, which proposed a very similar approach to Dense\-Net \cite{Huang2017a} with fully convolutional networks, with the main difference that the concatenation of feature maps between the encoding paths and decode paths was omitted. 
This means that, while in the work of \cite{Jegou2017a} only the last feature map of the convolutional layer is fed into the transition layer, Zhu et al.~\cite{Zhu2018a} propose to keep all feature maps and concatenate it before passing it to the transition layer.
It also avoids connection skipping due to weak correspondence and no max-pooling in encoding layers was used. To compensate for this, a stride of two was used. 
 
\begin{figure}[h!]
    \centering
    \includegraphics[scale=1.0]{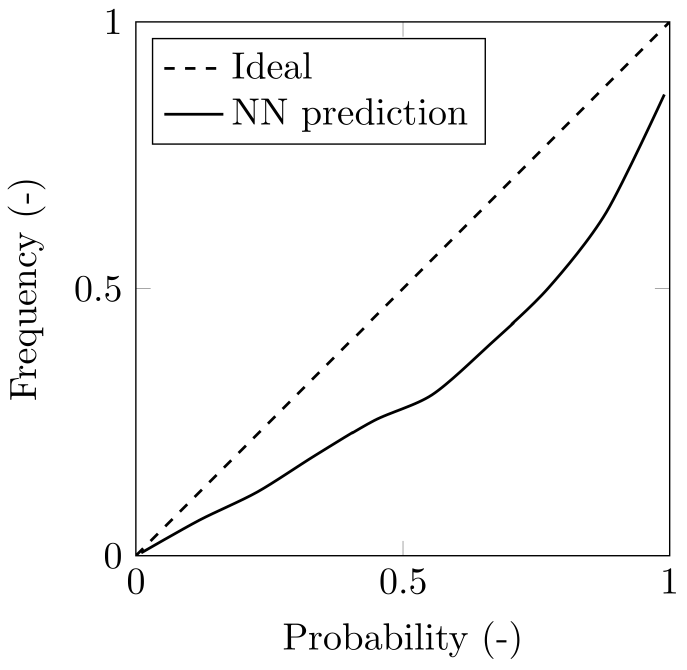}
    \vspace*{-0.35cm}
    \caption{The reliability plot illustrates the NN prediction or surrogate model that compares the NN prediction to the perfect (ideal) reliability line. The model frequency was evaluated at $30$ points with a maximum of 86\,$\%$.}
\label{fig:reliability-diagram}
\end{figure}

The overall loss accuracy of the constitutive model was $86\,$\%. The reliability plot is shown in Fig.~\ref{fig:reliability-diagram}.

\subsection{Training}
From a variety of options \cite{Ruoyu2020a,Soydaner2020a} we have chosen the optimizer ADAM \cite{Kingma2015a} to adjust the learning rate $\varepsilon_t$ in each iteration $t$ in (\ref{eq:w-update}).
Cosine annealing \cite{Loshchilov2017a} was selected as the learning rate schedule for $\varepsilon_t$ in (\ref{eq:w-update}), which resets the learning rate every $T_{\rm max}$ training epochs according to
\begin{equation}
    \varepsilon_t = \varepsilon_{\rm min} + \frac{1}{2} (\varepsilon_{\rm max}-\varepsilon_{\rm min}) \Big[1+\cos\Big(\frac{T_{\rm cur}}{T_{\rm max}}\pi\Big)\Big],
\end{equation}
where $\varepsilon_{\rm max}$ is the maximum learning rate set to the initial learning rate of $0.03$ and $\varepsilon_{\rm min}=0$ is the minimum learning rate, while $T_{\rm cur}$ denotes the number of epochs since the last restart and $T_{\rm max}=20$ denotes the maximum number of iterations. 
In other words, the scheduler resets the learning rate every $20$ epochs. A new epoch begins when the ADAM optimizer has gone through each batch of training data once, as the training data is partitioned into batches to trade computation time for GPU memory.
The gradient of the product in the posterior (\ref{eq:posterior-aux}) is then approximated as $\nabla_{\mathrm{w}} \prod_{n_\xi}^{N_\xi} \p{\obssmallset}{\xibf^{(n_\xi)},\wbf} \approx  \frac{N_\xi}{\vert V \vert} \nabla_{\mathrm{w}} \prod_{n_\xi \in V} \p{\obssmallset}{\xibf^{(n_\xi)}, \wbf}$, where $\vert V \vert$ denotes the number of indices of the subset of data in batch $V$.
The total number of epochs was $500$ with a constant batch size of $350$.
The training was performed on six Tesla K20Xm GPUs, each with 6\,GB GDDR5 memory. 

 \bibliographystyle{elsarticle-num} 
 \bibliography{Ranftl_et_al-JMPS-submitted}





\end{document}